\documentclass[reprint,aps,pre]{revtex4-1}

\usepackage{amsmath,amsfonts,amscd,amssymb,graphicx,amsthm}

\theoremstyle{definition}
\newtheorem{example}{Example}

\usepackage{subeqnar}
\usepackage{dcolumn}
\usepackage{booktabs}
\usepackage{overpic}
\usepackage{color}
\usepackage[caption=false]{subfig}
\usepackage{adjustbox}
\usepackage{rotating}
\usepackage{float}
\makeatletter
\let\newfloat\newfloat@ltx
\makeatother
\usepackage{algorithm}
\usepackage{algpseudocode}
\algnewcommand\algorithmicinput{\textbf{Input:}}
\algnewcommand\Input{\item[\algorithmicinput]}
\usepackage{placeins}

\begin{document}

\title{Reduced Order Modeling with Shallow Recurrent Decoder Networks}

\author{Matteo Tomasetto$^\dag$, Jan P. Williams$^\ddag$, Francesco Braghin$^\dag$, Andrea Manzoni$^*$, and J. Nathan Kutz$^{**}$}
\affiliation{$^\dag$ Department of Mechanical Engineering, Politecnico di Milano, Milano, Italy}
\affiliation{$^\ddag$ Department of Mechanical Engineering, University of Washington, Seattle, WA} 
\affiliation{$^{*}$ MOX - Department of Mathematics, Politecnico di Milano, Milano, Italy}
\affiliation{$^{**}$Department of Applied Mathematics and Electrical and Computer Engineering, University of Washington, Seattle, WA} 

\begin{abstract}
Reduced Order Modeling is of paramount importance for efficiently inferring high-dimensional spatio-temporal fields in parametric contexts, enabling computationally tractable parametric analyses, uncertainty quantification and control. However, conventional dimensionality reduction techniques are typically limited to known and constant parameters, inefficient for nonlinear and chaotic dynamics, and uninformed to the actual system behavior. In this work, we propose {\em sensor-driven} SHallow REcurrent Decoder networks for Reduced Order Modeling (SHRED-ROM). Specifically, we consider the composition of a long short-term memory network, which encodes the temporal dynamics of limited sensor data in multiple scenarios, and a shallow decoder, which reconstructs the corresponding high-dimensional states. SHRED-ROM is a robust {\em decoding-only} strategy that circumvents the numerically unstable approximation of an inverse which is required by encoding-decoding schemes.
To enhance computational efficiency and memory usage, the full-order state snapshots are reduced by, e.g., proper orthogonal decomposition, allowing for compressive training of the networks with minimal hyperparameter tuning. Through applications on chaotic and nonlinear fluid dynamics, we show that SHRED-ROM {\em (i)} accurately reconstructs the state dynamics for new parameter values starting from limited fixed or mobile sensors, independently on sensor placement, {\em (ii)} can cope with both physical, geometrical and time-dependent parametric dependencies, while being agnostic to their actual values, {\em (iii)} can accurately estimate unknown parameters, and {\em (iv)} can deal with different data sources, such as high-fidelity simulations, coupled fields and videos.
\end{abstract}
\maketitle


\section{INTRODUCTION} \label{intro}

\begin{figure*}[t]
    \centering
    \subfloat{
    \includegraphics[width=\textwidth]{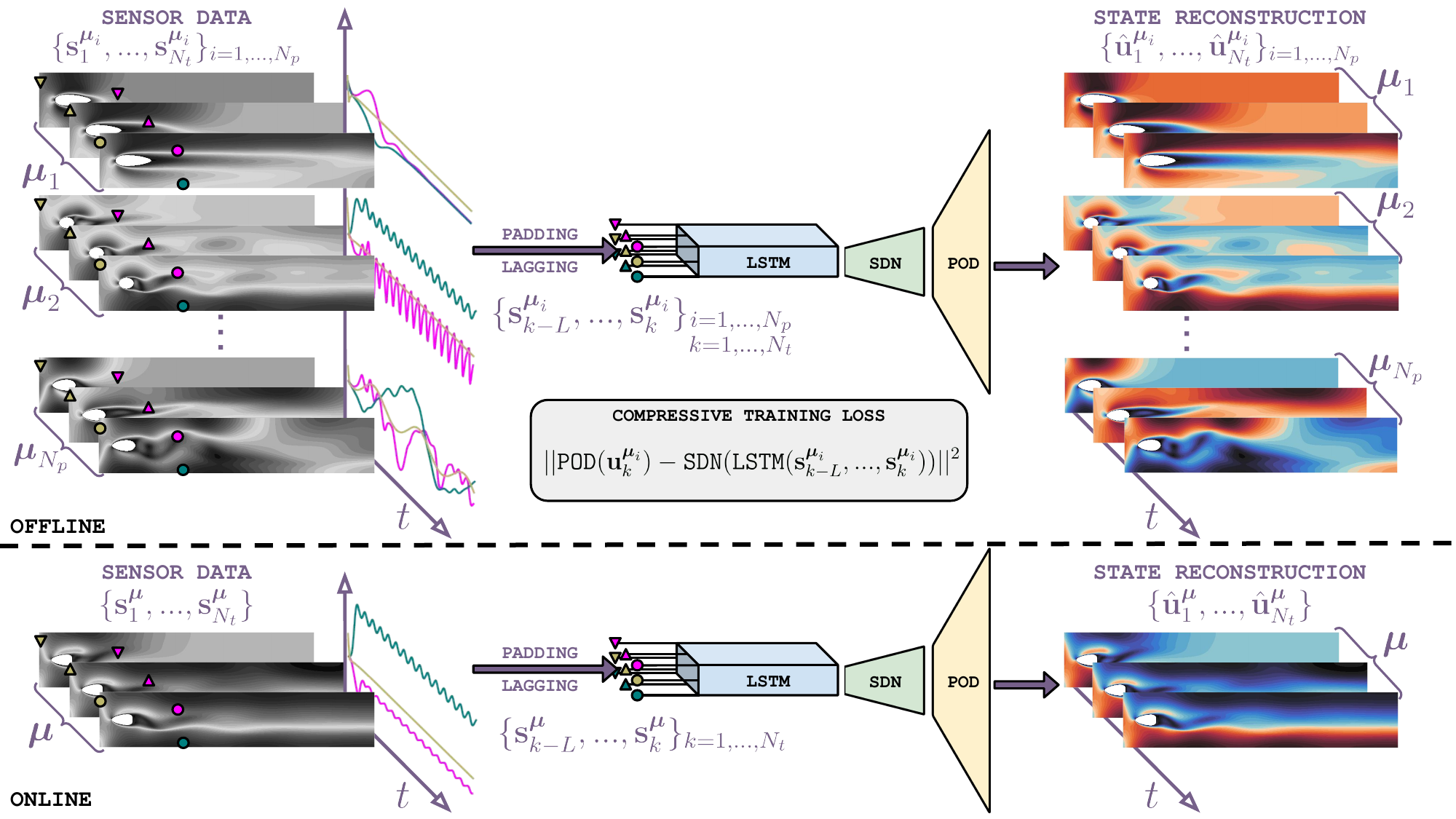}}
    \caption{Graphical summary of the {\em SHallow REcurrent Decoder-based Reduced Order Model} (SHRED-ROM). Sparse sensor values ${\bf s}_{k}^{\boldsymbol{\mu}_i}$ over time windows of length $L$ in multiple scenarios are encoded through a {\em long short-term memory} (LSTM), while a {\em shallow decoder network} (SDN) projects the resulting latent representation in the high-dimensional state space. Full-order state snapshots ${\bf u}_k^{\boldsymbol{\mu}_i}$ are reduced by {\em proper orthogonal decomposition} (POD), allowing for compressive training at the POD level. After training, in the online phase, it is possible to reconstruct high-dimensional state trajectories $\hat{{\bf u}}_k^{\boldsymbol{\mu}}$ for new parameters $\boldsymbol{\mu}$.}
    \label{fig:SHRED-ROM}
\end{figure*}

{\em Reduced order models} (ROMs) are widely used computational tools for accelerating engineering design and characterization~\cite{Benner2015siamreview,antoulas2005approximation,quarteroni2015reduced,hesthaven2016certified}. Specifically, scientific computing is now an integral part of every field of application, with high-fidelity numerical solvers and methods~\cite{kutz2013data} playing a critically enabling role in the modeling of high-dimensional, complex dynamical systems. However, this might be extremely demanding -- or even prohibitive -- in case repeated predictions are required for multiple scenarios, in very rapid times, or within sequential design or control 
loops. ROMs designate any approach aimed at replacing the high-fidelity problem by one featuring a much lower numerical complexity. In emerging modern applications, such as turbulence closure modeling, weather forecasting, powergrid networks, climate modeling and neuroscience, the construction of tractable dynamic models directly from data or high-fidelity simulations enables both engineering design and scientific discovery. Scientific machine learning~\cite{brunton2020data} is an emerging paradigm for constructing data-driven ROMs. As with traditional ROMs, machine learned ROMs aim to learn both low-dimensional embeddings and evolution dynamics which accurately reconstruct (in a least-square or statistical sense) the high-fidelity and high-dimensional state of the original (possibly parametric) system. In what follows, we advocate a new ROM architecture based upon the method of {\em separation of variables} for solving {\em partial differential equations} (PDEs). Specifically, we train a recurrent neural network to capture the temporal behavior of limited sensor data, while mapping its latent space to the high-dimensional state via a shallow decoder. The resulting {\em Shallow REcurrent Decoder-based Reduced Order Model} (SHRED-ROM) turns out to be an ultra-hyperreduced order modeling framework which provides a fully data-driven and robust ROM architecture for data or high-fidelity simulations. A graphical summary of the SHRED-ROM architecture for parametric state reconstruction from limited sensor data is presented in Figure~\ref{fig:SHRED-ROM}.

PDEs model a diversity of spatio-temporal systems, including those found in the classical physics fields of fluid dynamics, electrodynamics, heat conduction, and quantum mechanics.   Indeed, PDEs provide a canonical description of how a system evolves in space and time due the presence of partial derivatives which model the relationships between rates of change of time and space. Governing equations of physics-based systems simply provide a constraint, or restriction, on how these evolving quantities are related.  We consider PDEs of the form~\cite{courant2008methods}
\begin{equation}
  \dot{{u}} = N({u},{\bf x},t,\boldsymbol{\mu})
  \label{eq:PDE}
\end{equation}
    where ${u}({\bf x},t)$ is the variable of interest, or the state-space, which we are trying to model.  Alternatively, it may be that $N(\cdot)$ is unknown, so a purely data-driven strategy must be considered.  The solution ${u}({\bf x},t)$ of the PDE depends upon the spatial variable {\bf x}, which can be in 1D, 2D or 3D, and time $t$.  Importantly, solutions can often depend on a vector of parameters $\boldsymbol{\mu}$, thus requiring a solution that can model parametric dependencies, i.e.  ${u}({\bf x},t, \boldsymbol{\mu})$.
Solutions of (\ref{eq:PDE}) are typically achieved through numerical simulation, unless $N(\cdot)$ is linear and constant coefficient so that analytic solutions are available. Asymptotic and perturbation methods can also offer analytically tractable solution methods~\cite{kutz2020advanced}.  In many modern applications, discretization of the evolution for ${u}({\bf x},t,\boldsymbol{\mu})$ results in a high-dimensional system for which computations are prohibitively expensive. The goal of building ROMs is to approximate the full simulations of (\ref{eq:PDE}) through inexpensive tractable computations.

Traditional ROMs generate a computational proxy of (\ref{eq:PDE}) by projecting the governing PDE into a new coordinate system~\cite{Benner2015siamreview,antoulas2005approximation,quarteroni2015reduced,hesthaven2016certified}. Coordinate transformations are one of the fundamental paradigms for producing solutions to PDEs~\cite{keener2018principles}. Specifically, ROMs produce coordinate transformations from the simulation data itself. Thus, if the governing PDE (\ref{eq:PDE}) is discretized so that
${u}({\bf x},t) \rightarrow {\bf u}_{k}={\bf u}(t_k) \in \mathbb{R}^{N_h}$ for $k=1,\ldots,N_t$, then snapshots of the solution can be collected into the data matrix
\begin{equation*}
  {\bf X} =\left[ \begin{array}{cccc} | & | & \cdots & | \\
         {\bf u}_1 & {\bf u}_2 & \cdots & {\bf u}_{N_t} \\
         | & | & \cdots & |   \end{array} \right]
\end{equation*}
where ${\bf X}\in\mathbb{C}^{N_h\times N_t}$. In case of parametric dependencies, the data matrix contains the discretized states ${u}({\bf x},t, \boldsymbol{\mu}) \rightarrow {\bf u}_{k}^{\boldsymbol{\mu}}={\bf u}(t_k, \boldsymbol{\mu}) \in \mathbb{R}^{N_h}$ for $N_p$ different parametric instances, that is
\begin{equation*}
  {\bf X} =\left[ \begin{array}{ccccc} | & | & \cdots & | & | \\
         {\bf u}_{1}^{\boldsymbol{\mu}_1} & {\bf u}_{2}^{\boldsymbol{\mu}_1} & \cdots & {\bf u}_{N_t - 1}^{\boldsymbol{\mu}_{N_p}}  &  {\bf u}_{N_t}^{\boldsymbol{\mu}_{N_p}} \\
         | & | & \cdots & | & |   \end{array} \right] 
\end{equation*}
where ${\bf X}\in\mathbb{C}^{N_h\times N_t N_p}$. An optimal coordinate system for ROMs is extracted from the data matrix ${\bf X}$ with a {\em singular value decomposition} (SVD)~\cite{kutz2013data}:
\begin{equation*}
  {\bf X} = {\bf \boldsymbol{\Psi} \Sigma V}^*
\end{equation*}
where ${\boldsymbol{\Psi}}\in\mathbb{C}^{N_h\times r}$, ${\bf \Sigma}\in\mathbb{R}^{r\times r}$ and ${\bf V}\in\mathbb{C}^{N_t N_p\times r}$, with $N_p = 1$ for nonparametric problems, and $r$ is the number of modes extracted to represent the data. The SVD, which is more commonly known in the ROMs community as the {\em proper orthogonal decomposition} (POD)~\cite{holmes2012turbulence,Benner2015siamreview}, computes the dominant correlated activity of the data as a set of orthogonal modes. It is guaranteed to provide the best $\ell_2$-norm reconstruction of the data matrix ${\bf X}$ for a given number of modes $r$.  Importantly, the $r$ modes ${\boldsymbol{\Psi}}$ extracted from the data matrix are used to produce a separation of variables solution to the PDE~\cite{haberman1983elementary}:
\begin{equation}
  {\bf u} = {\boldsymbol{\Psi}} ({\bf x} ) {\bf a}(t, \boldsymbol{\mu})
  \label{eq:gal}
 \end{equation}
where the vector ${\bf a}={\bf a}(t,\boldsymbol{\mu})$ is determined by using this solution form and performing a Galerkin projection of (\ref{eq:PDE})~\cite{Benner2015siamreview}. Thus, a projection-based ROM simply represents the governing PDE evolution (\ref{eq:PDE}) in the $r$-rank subspace spanned by ${\boldsymbol{\Psi}}$.

The projection-based ROM construction often produces a low-rank model for the dynamics of ${\bf a}(t,\boldsymbol{\mu})$ that can be unstable~\cite{carlberg2017galerkin}. Machine learning techniques offer a diversity of alternative methods for computing the dynamics in the low-rank subspace. The simplest architecture aims to train a deep neural network that maps the solution forward in time, possibly being robust on different admissible scenario parameters:
\begin{equation}
   {\bf a}^{\boldsymbol{\mu}}_{k+1} = {\bf f}_{\boldsymbol{\theta}} ({\bf a}^{\boldsymbol{\mu}}_k)
   \label{eq:ak}
\end{equation}
where ${\bf a}^{\boldsymbol{\mu}}_k={\bf a}(t_k, \boldsymbol{\mu})$ and ${\bf f}_{\boldsymbol{\theta}}$ represents a deep neural network whose weights and biases $\boldsymbol{\theta}$ are determined by minimizing the prediction error on the available reduced snapshots ${\bf a}^{\boldsymbol{\mu}_i}_k$ for $i=1,\ldots,N_p$ and $k=1,\ldots,N_t$. A diversity of neural networks can be used to advance solutions, or learn the flow map from time $t$ to $t+\Delta t$~\cite{qin2019data,liu2020hierarchical}. Indeed, deep learning algorithms provide a flexible framework for constructing a mapping between successive time steps in multiple scenarios. Recently, Parish and Carlberg~\cite{parish2020time} developed a suite of neural network-based methods for learning time-stepping models for (\ref{eq:ak}), along with an extensive comparison between different neural network architectures and traditional techniques for time-series modeling.

Machine learning algorithms offer options beyond POD-Galerkin projection and deep learning-based modeling of the time-stepping in the variable ${\bf a}(t,\boldsymbol{\mu})$. Thus, instead of inserting (\ref{eq:gal}) back into (\ref{eq:PDE}) or learning a flow map ${\bf f}_{\boldsymbol{\theta}}$ for (\ref{eq:ak}), we can instead think about directly building a model for the dynamics of ${\bf a}(t,\boldsymbol{\mu})$, that is
\begin{equation*}
  \dot{{\bf a}} = {\bf f} ({\bf a},t,\boldsymbol{\mu})
\end{equation*}
where ${\bf f}(\cdot)$ now prescribes the dynamics. Two highly successful options have emerged for producing a  model for this dynamical system: (i) the {\em dynamic mode decomposition} (DMD)~\cite{Kutz2016book} and (ii) the {\em sparse identification of nonlinear dynamics} (SINDy)~\cite{Brunton2016pnas}. The DMD model for ${\bf f}(\cdot)$ is assumed to be linear so that
\begin{equation*}
    \dot{{\bf a}} \approx {\bf L} {\bf a}
\end{equation*}
where ${\bf L}$ is a linear operator found by regression. Solutions are then trivial as all that one requires is to find the eigenvalues and eigenvectors of ${\bf L}$ in order to provide a general solution by linear superposition. Note that, in parametric settings, it may be convenient to consider interpolation strategies among a set of linear operators ${\bf L}(\boldsymbol{\mu}_i)$ for $i=1,\ldots,N_p$, as proposed by Andreuzzi et al.~\cite{andreuzzi2023} and Huhn et al.~\cite{hhun2023}. The SINDy method makes, instead, a different assumption: the dynamics ${\bf f}(\cdot)$ can be represented by only a few terms. In this case, the regression has the form
\begin{equation*}
  \dot{{\bf a}} \approx \boldsymbol{\Theta}({\bf a},\boldsymbol{\mu}) \boldsymbol{\xi}
\end{equation*}
where the columns of the matrix $\boldsymbol{\Theta}({\bf a},\boldsymbol{\mu})$ are (possibly $\boldsymbol{\mu}$-dependent) candidate terms from which to construct a dynamical system and $\boldsymbol{\xi}$ contains the loading (or coefficient or weight) of each library term. SINDy assumes that $\boldsymbol{\xi}$ is a sparse vector so that most of the library terms do not contribute to the dynamics. The regression is a simple solve of an overdetermined linear system that is regularized by sparsity, or the sparsity-promoting $\ell_0$ or $\ell_1$ norms.  

In addition to the diversity of methods for building a model for the time dynamics of ${\bf a}(t,\boldsymbol{\mu})$, there also exists the possibility of using coordinates other than those defined by $\boldsymbol{\Psi}$. Moving beyond a linear subspace can provide improved coordinate systems for building latent dynamic models. Importantly, there exists the possibility of learning a coordinate system where, for instance, a linear DMD model or a parsimonious SINDy model can be more efficiently imposed. Thus we wish to learn a coordinate transformation 
\begin{equation*}
  {\bf z} = {\bf f}_{\boldsymbol{\theta}} ({\bf u})
\end{equation*}
where ${\bf z}$ is the new variable representing the state space ${\bf u}$ and ${\bf f}_{\boldsymbol{\theta}}$ is a neural network -- such as, e.g., the encoder part of an autoencoder -- that defines the coordinate transformation. This allows us to find a coordinate system beyond the standard linear, low-rank subspace $\boldsymbol{\Psi}$, which can be advantageous for ROM construction. The goal is then to fit a dynamical model in the new coordinate system
\begin{equation*}
\dot{{\bf z}} = {\bf f}({\bf z}, t, \boldsymbol{\mu})
\end{equation*}
through, for instance, a DMD or SINDy approximation \cite{Champion2019pnas}. A more recent extension to parametric problems has been proposed by Conti et al.~\cite{conti2023reduced}, also featuring uncertainty quantification by considering variational autoencoders \cite{conti2024veni}.
Similarly, Regazzoni et al.~\cite{regazzoni2019machine, Regazzoni2024} automatically discover a suitable low-dimensional coordinate system and the corresponding latent dynamics that best describe the input-output data pairs. A more flexible and theoretically rigorous formulation of the problem of latent dynamics learning has been proposed by Farenga et al.~\cite{farenga2025latent}, whereas alternative approaches to evolve the latent space dynamics using recurrent neural networks or an auto-regressive attention mechanism have been proposed by Koumoutsakos and coauthors \cite{vlachas2022multiscale,gao2024generative}.
Importantly, in contrast to autoencoders, the decoding-only strategy of SHRED-ROM does not require the computation of inverse pairs, i.e. an encoder and the corresponding decoder. Since the early days in scientific computing, it has been well known that the computation of the inverse of a matrix is highly unstable and non-robust~\cite{forsythe1977computer,croz1992stability,higham2002accuracy}. By decoding only, SHRED-ROM circumvents this problem and learns a single embedding without the corresponding inversion.

Instead of learning a dynamical model at POD or latent level, it is also possible to focus on the map from time-parameters to the POD coefficients $(t,\boldsymbol{\mu}) \to {\bf a}(t,\boldsymbol{\mu})$ or to other low-dimensional coordinate systems $(t,\boldsymbol{\mu}) \to {\bf z}(t,\boldsymbol{\mu})$. This is the case, for instance, of the non-intrusive reduced order modeling frameworks proposed by Hesthaven and Ubbiali~\cite{hesthaven2018} and Fresca et al.~\cite{fresca2022, Fresca2021}, where long-short term memory networks have also been employed \cite{fresca2023long}.

In what follows, we introduce a new reduced order modeling framework based upon the separation of variable method that combines recurrence and decoding in order to reconstruct high-dimensional states starting from limited sensor measurements, while being agnostic to sensor placement and parameter values. Differently from the aforementioned ROM strategies, we encode the sensor dynamics through a recurrent neural network, mapping its latent representation to the state space with a shallow decoder. The proposed decoding-only strategy turns out to be very efficient and accurate in a wide range of applications, such as the reconstruction of chaotic and nonlinear fluid dynamics in multiple scenarios.

\section{SHALLOW RECURRENT DECODER-BASED REDUCED ORDER MODELING}

The {\em shallow recurrent decoder network} (SHRED) proposed by Williams et al.~\cite{williams2024} is a promising sensing strategy, with impressive reconstruction and forecasting results in the low-data limit. See~\cite{kutz2024shallowrecurrentdecoderreduced, ebers2024, riva2024robuststateestimationpartial, gao2025sparseidentificationnonlineardynamics} for a complete presentation with possible extensions. In this work, we extend the SHRED strategy to a unified reduced order modeling framework capable of {\em (i)} reconstructing high-dimensional dynamics from sparse sensor measurements, regardless of sensor placement, {\em (ii)} dealing with both physical, geometrical and time-dependent parametric dependencies, while being agnostic to their actual values, {\em (iii)} estimating unknown parameters, and {\em (iv)} coping with both fixed or mobile sensors, as well as different data sources such as high-fidelity simulations, coupled fields or videos. Importantly, computational efficiency and memory usage are enhanced by reducing the dimensionality of full-order snapshots, allowing for compressive training of the networks.

\subsection{The rationale behind SHRED-ROM: separation of variables}

SHRED-ROM consists of a combination of recurrence and decoding, as better detailed in Section~\ref{subsec:SHRED-ROM_model}. While the former operation captures the temporal behavior of limited sensor data, the latter performs a spatial upscaling to recover the corresponding high-dimensional state. A similar split in spatial and temporal components is taken into account by the well-known method of separation of variables for solving linear PDEs~\cite{folland1995introduction}, which assumes that the solution has the form $u({\bf x},t,\boldsymbol{\mu})= T(t,\boldsymbol{\mu}) X({\bf x},\boldsymbol{\mu})$. For the sake of simplicity, we first review the case of nonparametric problems, postponing at the end of the section the case of parametric PDEs.
\vspace{0.4 cm}

\textbf{Linear PDEs.} Let us consider the 1D constant coefficient linear PDE
\begin{equation}
 \dot{u} = {\cal L} {u}
 {\label{eq:linearPDE}}
\end{equation}
where ${\cal L}={\cal L}(\partial_x, \partial^2_x, \ldots) $ is a linear differential operator modeling the dynamics of $u(x,t)$. Note that (\ref{eq:linearPDE}) has to be coupled with suitable initial and boundary conditions in order to guarantee its well-posedness. When looking for solutions in the separated form $u(x,t)= T(t) X(x)$, (\ref{eq:linearPDE}) reduces to two differential equations for, respectively, time $T(t)$ and space $X(x)$, namely
\begin{equation}
 \dot{T} = \lambda T; \qquad  {\cal L} {X} = \lambda X,
 \label{eq:eigenvalTX}
\end{equation}
where $\lambda \in \mathbb{C}$. Finally, thanks to the linearity assumption, the explicit formula for $u(x,t)$ is given by superposition of the solutions of (\ref{eq:eigenvalTX}), that is
\begin{equation}
    u(x,t) = \sum_{n=1}^{N} a_n \exp(\lambda_n t) \phi_n(x)
    \label{eq:ef}
\end{equation}
where $\phi_n(x)$ and $\lambda_n$ are, respectively, the eigenfunctions and eigenvalues of the linear operator ${\cal L}$, i.e. ${\cal L}\phi_n(x) = \lambda_n \phi_n(x)$. Note that the sum in (\ref{eq:ef}) is truncated up to $N$ terms, as standard practice for numerical evaluation. To uniquely determine the coefficients $a_n$, the initial condition $u(x,0)=u_0(x)$ is typically imposed in (\ref{eq:ef}), while exploiting the orthogonality of the eigenfunctions $\phi_n(x)$, yielding
\begin{equation*}
    u_0(x) = \sum_{n=1}^{N} a_n  \phi_n(x) \implies a_n =\langle u_0(x), \phi_n(x) \rangle
\end{equation*}
where $\langle\cdot,\cdot\rangle$ stands for the inner product.

In general, when dealing with real-world high-dimensional problems, it is more likely to measure the dynamical system in few locations over time. Therefore, the coefficients $a_n$ can be determined requiring that the solution ($\ref{eq:ef}$) matches the available sensor measurements. For instance, if $N$ temporal observations of a single sensor located in $x=x_s$ are available, the system of equations
\begin{equation*}
    u(x_s, t_k) = \sum_{n=1}^{N} a_n 
    \exp(\lambda_n t_k) \phi_n(x_s) 
    \,\,\,\,\,\, \mbox{for} \,\,\, k=1,\ldots,N
\end{equation*}
uniquely determine the solution $u(x,t)$. Similarly, the high-dimensional state may be prescribed by employing multiple sensors over time. For example, if two sensors are available in the considered domain, then $N/2$ temporal observations steps are needed to compute the coefficients $a_n$. In general, we can exploit $N_s$ sensors on $N/N_s$ trajectory points. 

Besides the initial condition and the temporal history of stationary sensors, mobile sensors can be exploited, as shown in Section~\ref{subsec:SWE} and Section~\ref{subsec:pinball} with sensors passively transported by the underlying fluid. For example, if we take into account a single mobile sensor with time-dependent position $x_s = x_s(t)$, the constraints become
\begin{equation*}
    u(x_s(t_k), t_k) = \sum_{n=1}^{N} a_n 
    \exp(\lambda_n t_k) \phi_n(x_s(t_k))
\end{equation*}
for $k=1,\ldots,N$. The temporal trajectory information at a single spatial location, as well as a mobile sensor, is therefore equivalent to knowing the full spatial field at a given time instant. The above arguments guarantee that, in the linear case, SHRED-ROM reconstructs exactly the high-dimensional spatio-temporal state. Note that the same argument may be easily extended to systems in higher spatial dimensions.
\vspace{0.4 cm}

\textbf{Nonlinear PDEs.} Let us consider the 1D nonlinear PDE
\begin{equation}
 \dot{u} = N(u)
 {\label{eq:nonlinearPDE}}
\end{equation}
equipped with suitable initial and boundary conditions, where $N = N(\partial_x, \partial^2_{x},\ldots)$ is a nonlinear differential operator. The dynamical system in (\ref{eq:nonlinearPDE}) is typically solved through numerical techniques such as, e.g., finite differences, finite element methods, finite volume methods and spectral methods~\cite{kutz2013data}. In the latter case, the solution is approximated by a spectral basis expansion
\begin{equation}
    u(x,t) = \sum_{n=1}^{N} a_n (t) \phi_n(x).
    \label{eq:spectral}
\end{equation}
Standard choices of spectral basis functions $\{{\phi_n(x)}\}_{n=1}^{N}$ include, e.g., Fourier modes or Chebyshev polynomials. Inserting (\ref{eq:spectral}) back into (\ref{eq:nonlinearPDE}) yields a system of $N$ coupled ODEs for $a_n(t)$:
\begin{equation}
  \dot{a}_n=f_n (a_1, a_2, \cdots, a_N) \,\,\,\,\,\, \mbox{for} \,\,\, n=1,\ldots, N .
  \label{eq:an}
\end{equation}
Similarly to the method of separation of variables in the linear limit, the solution of this $N$-dimensional ODE depends on $N$ constants of integration to be determined by imposing initial condition and orthogonality in (\ref{eq:spectral}), that is
\begin{equation*}
    a_n(0)= \langle u_0(x), \phi_n(x) \rangle . 
\end{equation*}
If a fixed sensor monitors the PDE solution in $x=x_s$ over $N$ time steps, we can still uniquely determine the dynamics of the time-dependent coefficients $a_n$ in (\ref{eq:an}). Specifically, we impose the constraints
\begin{equation*}
    u(x_s,t_k)=\sum_{n=1}^N a_n(t_k)\phi_n(x_s)  \,\,\,\,\,\, \mbox{for} \,\,\, k=1,\ldots, N
\end{equation*}
to determine the $N$ constants of integration that arise while solving (\ref{eq:an}), thus obtaining the state approximation through (\ref{eq:spectral}). In the nonlinear case, establishing rigorous theoretical bounds for SHRED-ROM poses challenges, consistently with the difficulties encountered to rigorously bound both analytical and numerical solutions in computational PDE settings. However, time-series of sparse sensor measurements allows us to recover standard numerical techniques, such as the spectral method. The same argument can be then extended to multiple or mobile sensors, as detailed for the linear case. 
\vspace{0.4 cm}

\textbf{Coupled PDEs.} Let us now consider coupled, constant coefficient 1D linear PDEs, such as the first-order coupled system
\begin{subeqnarray}
 \dot{u} = {\cal L}_1  u + {\cal L}_2 v
 \\
 \dot{v} = {\cal L}_3  u + {\cal L}_4 v
 {\label{eq:linearPDE2}}
\end{subeqnarray}
in the unknowns $u(x,t)$ and $v(x,t)$, where ${\cal L}_i = {\cal L}_i(\partial_x,\partial^2_x,\ldots)$ for $i=1,2,3,4$ are linear differential operators. If we differentiate (\ref{eq:linearPDE2}a) with respect to time and substitute ${v}$ and $\dot{v}$ with, respectively, (\ref{eq:linearPDE2}a) and (\ref{eq:linearPDE2}b), we rewrite the coupled system as a second-order PDE
\begin{equation*}
    \ddot{u} = {\cal L}_1 \dot{u}
    +{\cal L}_2 {\cal L}_3 u
    +{\cal L}_2 {\cal L}_4 \left(
     {\cal L}_2^{-1} (\dot{u} - {\cal L}_1 {u})
    \right)
\end{equation*}
dependent only on the spatio-temporal field $u(x,t)$. This suggests that the solution fields $u(x,t)$ and $v(x,t)$ can be reconstructed by only knowing $u(x,t)$. For instance, in the application detailed in Section~\ref{subsec:SWE}, we reconstruct the high-dimensional flow velocity exploiting sensors monitoring a coupled quantity, that is the height deviation of the pressure surface from its mean height. Second-order PDEs require both an initial condition $u(x,0)$ and an initial velocity $\dot{u}(x,0)$ to uniquely determine the solution. As with previous arguments, we can show that the high-dimensional coupled fields can be prescribed by, e.g., a single sensor measurement over $2N$ temporal trajectory points, as well as by multiple or mobile sensors.
\vspace{0.4 cm}

\textbf{Parametric PDEs.} Let us consider the parametric, constant coefficient 1D linear PDE
\begin{equation}
 \dot{u} = {\cal L}^{\boldsymbol{\mu}}{u}
 \label{eq:linearPDEparam}
\end{equation}
with suitable initial and boundary conditions, where ${\cal L}^{\boldsymbol{\mu}}={\cal L}(\partial_x, \partial^2_x, \ldots, {\boldsymbol{\mu}})$ is a parameters-dependent linear differential operator. As with the linear nonparametric case, looking for solutions of (\ref{eq:linearPDEparam}) in the separated form $u(x,t,{\boldsymbol{\mu}})= T(t,{\boldsymbol{\mu}}) X(x,{\boldsymbol{\mu}})$ leads to the differential equations
\begin{equation*}
 \dot{T} = \lambda T; \qquad  {\cal L}^{{\boldsymbol{\mu}}} {X} = \lambda X,
\end{equation*}
with $\lambda \in \mathbb{C}$. The parametric high-dimensional state $u(x,t,\boldsymbol{\mu})$ is then given by
\begin{equation*}
    u(x,t, \boldsymbol{\mu}) = \sum_{n=1}^{N} a_n \exp(\lambda^{{\boldsymbol{\mu}}}_n t) \phi^{{\boldsymbol{\mu}}}_n(x)
\end{equation*}
where $\phi^{{\boldsymbol{\mu}}}_n(x)$ and $\lambda^{{\boldsymbol{\mu}}}_n$ are, respectively, the eigenfunctions and eigenvalues of the parametric linear operator ${\cal L}^{\boldsymbol{\mu}}$, i.e. ${\cal L}^{\boldsymbol{\mu}} \phi^{\boldsymbol{\mu}}_n(x) = \lambda^{\boldsymbol{\mu}}_n \phi^{{\boldsymbol{\mu}}}_n(x)$. As with previous arguments, the coefficients $a_n$ and thus the solution $u(x,t,\boldsymbol{\mu})$ can be uniquely determined by employing a fixed sensor in position $x = x_s$ monitoring the state on $N$ trajectory points in the scenario identified by the set of parameters $\boldsymbol{\mu}$, that is
\begin{equation*}
    u(x_s, t_k, \boldsymbol{\mu}) = \sum_{n=1}^{N} a_n 
    \exp(\lambda^{\boldsymbol{\mu}}_n t_k) \phi^{\boldsymbol{\mu}}_n(x_s) 
    \,\,\,\,\,\, \mbox{for} \,\,\, k=1,\ldots,N .
\end{equation*}
The same arguments can be generalized to multiple or mobile sensors, to nonlinear and coupled parametric PDEs, as detailed in the previous cases, as well as to systems in higher spatial dimensions or {\em ordinary differential equations} (ODEs), as highlighted in the following example.

\begin{example} Let us consider the $3 \times 3$ parametric linear system of ODEs
\begin{equation}
\dot{{\bf u}} =
\begin{bmatrix}
1 & 2 & 0 \\ 0 & -1 & \mu \\ 0 & 0 & 2
\end{bmatrix} 
{\bf u}
\label{eq:ODE}
\end{equation}
where ${\bf u}(t,\mu) = [u_1(t,\mu), u_2(t,\mu), u_3(t,\mu)]^{\top}$. The explicit solution of (\ref{eq:ODE}) is given by
\begin{equation*}
\begin{bmatrix}
u_1(t,\mu) \\ u_2(t,\mu) \\ u_3(t,\mu)
\end{bmatrix} 
= c_1e^{2t}
\begin{bmatrix}
2\mu \\ \mu \\ 3
\end{bmatrix} 
+ c_2e^{- t}
\begin{bmatrix}
-1 \\ 1 \\ 0
\end{bmatrix} 
+ c_3e^{t}
\begin{bmatrix}
1 \\ 0 \\ 0
\end{bmatrix} 
\end{equation*}
where the coefficients $c_1, c_2, c_3$ has to be determined. To this aim, we may exploit the initial condition ${\bf u}(0,\mu) = [1, 2, 3]^{\top}$, that is we solve the system
\begin{equation*}
\begin{bmatrix}
u_1(0,\mu) \\ u_2(0,\mu) \\ u_3(0,\mu)
\end{bmatrix} 
= \begin{bmatrix}
1 \\ 2 \\ 3
\end{bmatrix} 
= 
\begin{bmatrix}
2\mu & -1 & 1 \\ \mu & 1 & 0 \\ 3 & 0 & 0
\end{bmatrix} 
\begin{bmatrix}
c_1 \\ c_2 \\ c_3
\end{bmatrix} 
\end{equation*}
yielding $c_1=1$, $c_2=2-\mu$ and $c_3=3-3\mu$.
Equivalently, we can determine the coefficients by employing one sensor monitoring $u_1$ over time. Specifically, if we measure $[u_1(0,1), u_1(1,1), u_1(2,1)]^{\top}=[1,e^2-e^{-1}, 2e^4-e^{-2}]$ in the scenario identified by $\mu=1$, we obtain the same results by solving the system
\begin{equation*}
\begin{bmatrix}
u_1(0,1) \\ u_1(1,1) \\ u_1(2,1)
\end{bmatrix} 
= \begin{bmatrix}
1 \\ 2e^{2}-e^{-1} \\ 2 e^{4} - e^{-2}
\end{bmatrix} 
= \begin{bmatrix}
2 & -1 & 1  \\ 2 e^{2} & -e^{-1} & e  \\ 2 e^{4} & -e^{-2} & e^2
\end{bmatrix} 
\begin{bmatrix}
c_1 \\ c_2 \\ c_3
\end{bmatrix} 
\end{equation*}
The same strategy can be then repeated for different values of the parameter $\mu$. Note that, if we consider measurements of $u_2$ and/or $u_3$ only, the coefficient $c_3$ does not appear in the corresponding system of constraints, and it is not possible to uniquely determine the exact solution. This limitation is due to the non-observability of the system with respect to $u_2$ and/or $u_3$. Thus, in this setting, full state reconstruction from sensor data is guaranteed whenever we have observability with respect to the measured quantity. 
\end{example}


\subsection{SHRED-ROM architecture and compressive training}
\label{subsec:SHRED-ROM_model}

SHRED-ROM aims to reconstruct a high-dimensional spatio-temporal field starting from $N_s$ sensors in multiple scenarios. Rather than learning the {\em one-shot} reconstruction map ${\bf s}_k^{\boldsymbol{\mu}} = \{u({\bf x}_s, t_k, \boldsymbol{\mu})\}_{s=1}^{N_s} \to {\bf u}_k^{\boldsymbol{\mu}}$ for $k=1,\ldots,N_t$ -- as considered by, e.g.,~\cite{nair2020, maulik2020, luo2023, zhang2025} -- we take advantage of the temporal history of sensor values. Specifically, a recurrent neural network ${\bf f}_T$ encodes the sensor measurements over a time window of length $L \leq N_t$ into a latent representation of dimension $N_l$, that is
\begin{align*}
    &{\bf f}_T:\underset{L \text{ times }}{\underbrace{\mathbb{R}^{N_s} \times \ldots \times \mathbb{R}^{N_s}}} \to \mathbb{R}^{N_l},
    \\
    &{\bf h}_k^{\boldsymbol{\mu}} = {\bf {f}}_T({\bf s}_{k-L}^{\boldsymbol{\mu}}, \ldots, {\bf s}_k^{\boldsymbol{\mu}}),
\end{align*}
where the hyperparameter $L$ identifies the number of lags considered by SHRED-ROM, and may be selected according to the problem-specific evolution rate or periodicity. The high-dimensional state is then approximated through a decoder ${\bf f}_X$, which performs a nonlinear projection of the latent representation onto the state space
\begin{align*}
    &{\bf f}_X: \mathbb{R}^{N_l} \to \mathbb{R}^{N_h},
    \\
    &{\bf u}_k^{\boldsymbol{\mu}} \approx \hat{\bf u}_k^{\boldsymbol{\mu}} = {\bf f}_X({{\bf h}_k^{\boldsymbol{\mu}}}) = {\bf f}_X({\bf f}_T({\bf s}_{k-L}^{\boldsymbol{\mu}}, \ldots, {\bf s}_k^{\boldsymbol{\mu}})).
\end{align*}
In this work, we consider a {\em long short-term memory} (LSTM) network~\cite{hochreiter1997long} to model the temporal dependency of sensor data, and a {\em shallow decoder network} (SDN) as latent-to-state map. In general, SHRED-ROM provides a modular framework where different architectures -- such as, e.g., convolutional neural networks~\cite{lecun2015deep}, gated recurrent units~\cite{chung2014}, echo state networks~\cite{Lukosevicius2012}, transformers~\cite{vaswani2023} and variational autoencoders~\cite{Kingma_2019} -- may be exploited.

In the offline phase, once for all, the neural networks appearing in SHRED-ROM have to be properly trained. To this aim, suppose to collect snapshots of the high-dimensional state -- either in the form of experimental measurements or synthetic data generated by high-fidelity simulations -- along with the corresponding sensor data
\begin{align*}
    {\bf u}_k^{\boldsymbol{\mu}_i} &= {\bf u}(t_k, \boldsymbol{\mu}_i) \in \mathbb{R}^{N_h},
    \\
    {\bf s}_k^{\boldsymbol{\mu}_i} &= \{u({\bf x}_s, t_k, \boldsymbol{\mu}_i)\}_{s=1}^{N_s} \in \mathbb{R}^{N_s}
\end{align*}
for different time instants $t_1,\ldots,t_{N_t}$ and scenario parameters ${\boldsymbol{\mu}_1},\ldots,{\boldsymbol{\mu}_{N_p}}$ in the parameter space $\mathcal{P}$. Note that, as discussed in the previous section, mobile sensors can be easily taken into account, resulting in the measurements ${\bf s}_k^{\boldsymbol{\mu}_i} = \{u({\bf x}_s(t_k), t_k, \boldsymbol{\mu}_i)\}_{s=0}^{N_s} \in \mathbb{R}^{N_s}$. Note also that, in order to strengthen the generalization capabilities of SHRED-ROM, it is crucial to adequately explore the variability in time and in the parameter space. Starting from the sequences of sensor measurements in multiple scenarios, we extract time-series of length $L$
\begin{equation}
\begin{array}{cc}
    \{{\bf s}_{k-L}^{\boldsymbol{\mu}_i}, \ldots, {\bf s}_k^{\boldsymbol{\mu}_i} \} \,\,\,\,\,\, \mbox{for} \,\,\, &i=1,\ldots,N_p
    \\
    &k=1,\ldots,N_t
\label{eq:sensor_lag}
\end{array}
\end{equation}
where pre-padding (${\bf s}_k^{\boldsymbol{\mu}_i} = {\bf 0}$ for $k \leq L$ and $i = 1,\ldots,N_p$) is applied to obtain state reconstructions on the whole time interval $[t_1, t_{N_t}]$, avoiding any burn-in period. After a training-validation-test splitting of the available input-output pairs, the recurrent neural network ${\bf f}_X$ and the shallow decoder ${\bf f}_T$ are trained by minimizing the reconstruction error
\begin{equation}
\begin{aligned}
     J({\bf u}_k^{\boldsymbol{\mu}_i}, \hat{\bf u}_k^{\boldsymbol{\mu}_i})  &= \sum_{\substack{i \in I_{\text{train}} \\ k \in K_{\text{train}}}} ||{\bf u}_k^{\boldsymbol{\mu}_i} - \hat{\bf u}_k^{\boldsymbol{\mu}_i}||^2  \\
     &= \sum_{\substack{i \in I_{\text{train}} \\ k \in K_{\text{train}}}} ||{\bf u}_k^{\boldsymbol{\mu}_i} - {\bf f}_X({\bf {f}}_T({\bf s}_{k-L}^{\boldsymbol{\mu}_i}, \ldots, {\bf s}_k^{\boldsymbol{\mu}_i}) )||^2,
\end{aligned}
\end{equation}
where $I_{\text{train}}$ and $K_{\text{train}}$ identify the times and scenarios in the training set, while $||\cdot||$ stands for the Euclidean norm. Once trained, in the online phase, SHRED-ROM provides a real-time reconstruction of the state trajectory starting from the corresponding sensor history in new time instants and/or new scenarios $\boldsymbol{\mu}$ unseen during training. 

Whenever the state dimension $N_h$ is remarkably high, compressive training strategies may be employed to enhance computational efficiency and memory usage. Specifically, it is possible to reduce the state dimensionality through a data- or physics-driven basis expansion of the snapshots ${\bf u}_k^{\boldsymbol{\mu}_i}$. Doing so, SHRED-ROM estimates only the $r \ll N_h$ basis expansion coefficients, rather than the entire high-dimensional state. For example, we can project the state snapshots onto a lower-dimensional subspace by POD, that is ${\bf u}_k^{\boldsymbol{\mu}_i} = {\bf \Psi} {{\bf a}_k^{\boldsymbol{\mu}_i}}$, resulting in the SHRED reconstruction 
\begin{equation*}
    \hat{\bf u}_k^{\boldsymbol{\mu}_i} = {\bf \Psi}\hat{\bf a}_k^{\boldsymbol{\mu}_i} = {\bf \Psi}{\bf f}_X({\bf {f}}_T({\bf s}_{k-L}^{\boldsymbol{\mu}_i}, \ldots, {\bf s}_k^{\boldsymbol{\mu}_i})).
\end{equation*}
An example of an alternative basis is provided in Section~\ref{subsec:SWE}, where spherical harmonics are taken into account to properly describe and compress synthetic state snapshots simulated on a sphere.


\section{NUMERICAL RESULTS}
\label{sec:test}

This section presents five test cases where SHRED-ROM is employed to reconstruct high-dimensional spatio-temporal fields starting from limited sensor measurements. In Section~\ref{subsec:SWE}, we employ SHRED-ROM to reconstruct synthetic spherical snapshots in a nonparametric setting, while exploiting both fixed and mobile sensors, as well as different compressive training strategies based on POD and spherical harmonics. To demonstrate the wide applicability of SHRED-ROM, we cope with the reconstruction of videos given limited pixel values in Section~\ref{subsec:GoPro}. The remaining three test cases are devoted to the reconstruction of high-dimensional states of chaotic, nonlinear and parametric fluid dynamics in multiple scenarios. Note that, when dealing with nonparametric problems, $80\%$ of the time instants are randomly selected as training sequences in $K_{\text{train}}$, while the remaining ones are equally split in validation and test. Instead, in parametric contexts, we consider a parameter-wise splitting -- i.e. $80\%$ of the trajectories are regarded as training set, while the remaining trajectories related to different parameter values constitute validation and test sets.

Differently to many deep learning-based models, SHRED-ROM requires minimal hyperparameter tuning. Indeed, in all the test cases presented, the following lightweight architecture is exploited: the LSTM ${\bf f}_T$ shows $2$ hidden layers with $64$ neurons each, while the SDN ${\bf f}_X$ is made of $2$ hidden layers having $350$ and $400$ neurons, respectively. Moreover, we select ReLU as activation function, and we prevent overfitting through dropout with rate equal to $0.1$. Regarding the neural networks training, we exploit Adam optimizer for $200$ epochs, half with learning rate equal to $0.001$ and half with learning rate equal to $0.0001$, considering a batch size equal to $64$. The relatively light and simple SHRED-ROM architecture, along with compressive training strategies, allows us to efficiently solve a wide range of challenging reconstruction problems with laptop-level computing.

The generalization capabilities of SHRED-ROM on new times and new scenarios in the test set are quantitatively assessed with the following mean relative error
\begin{equation*}
\begin{aligned}
     \varepsilon({\bf u}_k^{\boldsymbol{\mu}_i}, \hat{\bf u}_k^{\boldsymbol{\mu}_i})  &= \dfrac{1}{N_{\text{test}}}\sum_{\substack{i \in I_{\text{test}} \\ k \in K_{\text{test}}}} \dfrac{||{\bf u}_k^{\boldsymbol{\mu}_i} - \hat{\bf u}_k^{\boldsymbol{\mu}_i}||}{||{\bf u}_k^{\boldsymbol{\mu}_i}||}  \\
     &= \dfrac{1}{N_{\text{test}}}\sum_{\substack{i \in I_{\text{test}} \\ k \in K_{\text{test}}}} \frac{||{\bf u}_k^{\boldsymbol{\mu}_i} - {\bf f}_X({\bf {f}}_T({\bf s}_{k-L}^{\boldsymbol{\mu}_i}, \ldots, {\bf s}_k^{\boldsymbol{\mu}_i}) )||}{||{\bf u}_k^{\boldsymbol{\mu}_i}||},
\end{aligned}
\end{equation*}
where $I_{\text{test}}$ and $K_{\text{test}}$ identify the time and parameters in the test set, while $N_{\text{test}}$ is the test set cardinality.

\subsection{Shallow Water Equations}
\label{subsec:SWE}

\begin{figure*}
    \centering
    \begin{sideways}
    \makebox[0pt][l]{\hspace{-3.75cm}
    \begin{minipage}{4cm}
    \hphantom{} \\ {\bf \scriptsize DATA} \\   \hphantom{} \end{minipage}}
    \end{sideways}\subfloat[\shortstack{Fixed height sensors \\ Height reconstruction \\ 444 hours}]{
        \includegraphics[width=0.19\textwidth]{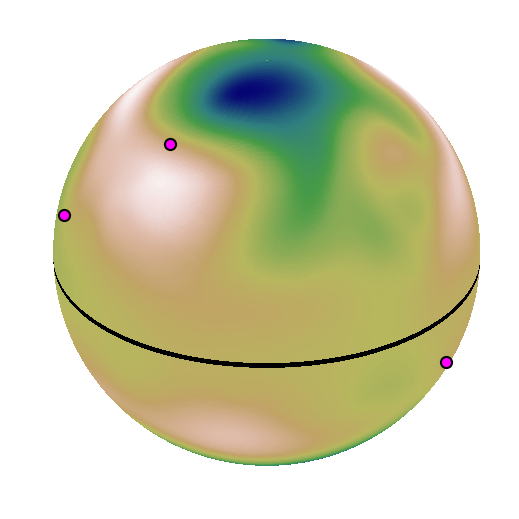}
      
    }\quad
    \subfloat[\shortstack{Fixed height sensors \\ Velocity reconstruction \\ 444 hours}]{
        \includegraphics[width=0.19\textwidth]{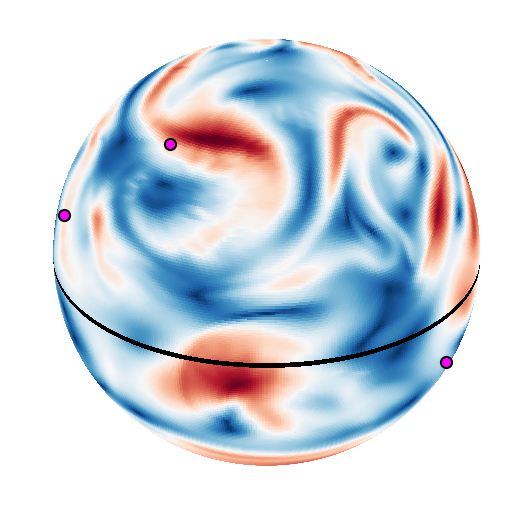}
      
    }\quad
    \subfloat[\shortstack{Mobile sensor \\ Velocity reconstruction \\ 11 hours}]{
        \includegraphics[width=0.19\textwidth]{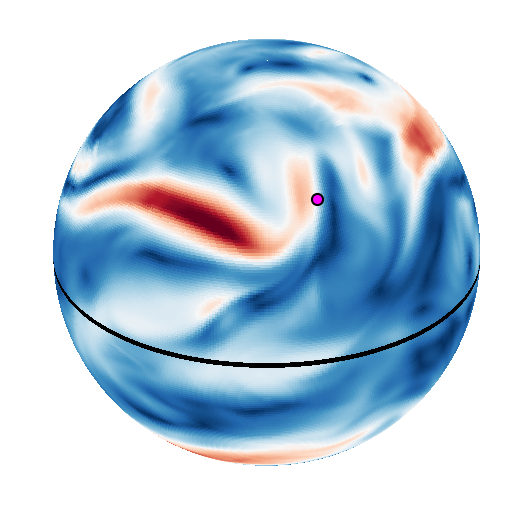}
      
    }\quad
    \subfloat[\shortstack{Mobile sensor \\ Velocity reconstruction \\ 957 hours}]{
        \includegraphics[width=0.19\textwidth]{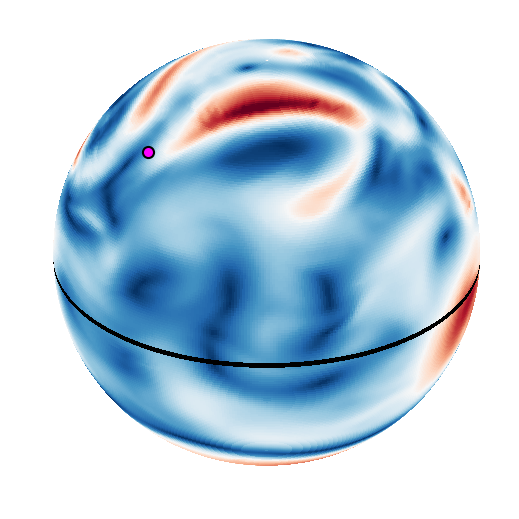}
      
    }

    \begin{sideways}
    \makebox[0pt][l]{\hspace{-3.75cm}
    \begin{minipage}{4cm}
    {\bf \scriptsize SHRED-ROM} \\   {\bf \scriptsize   WITH POD} \end{minipage}}
    \end{sideways}\subfloat{
        \includegraphics[width=0.19\textwidth]{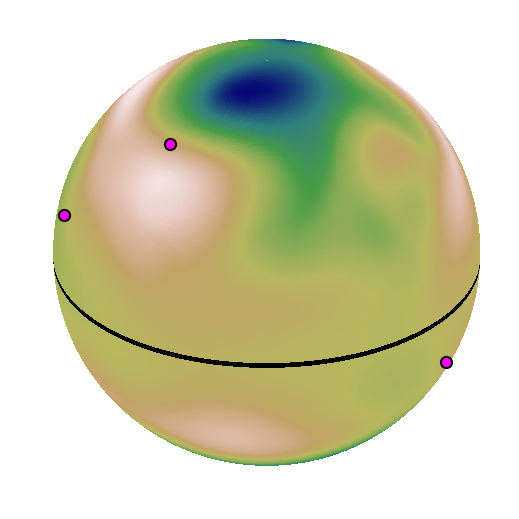}
      
    }\quad
    \subfloat{
        \includegraphics[width=0.19\textwidth]{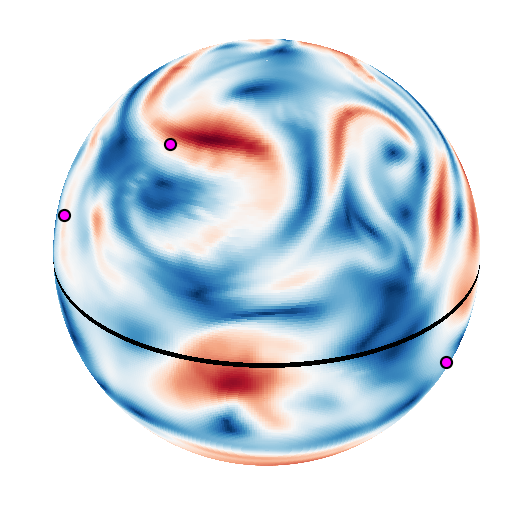}
      
    }\quad
    \subfloat{
        \includegraphics[width=0.19\textwidth]{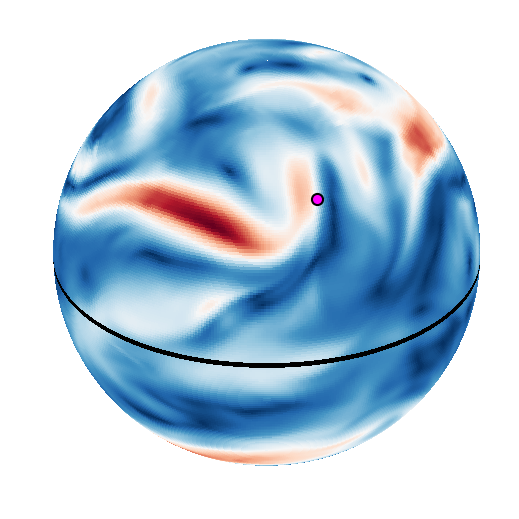}
      
    }\quad
    \subfloat{
        \includegraphics[width=0.19\textwidth]{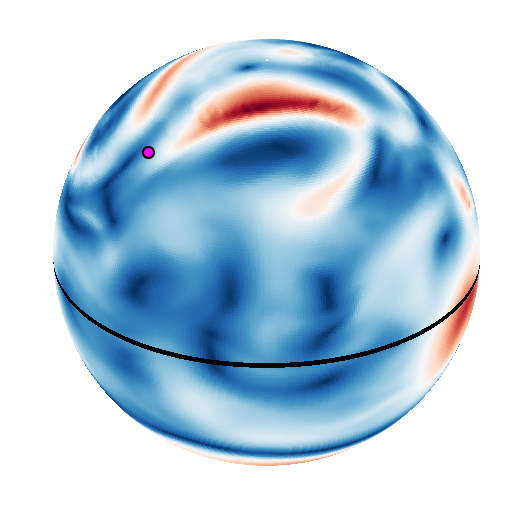}
      
    }

    \begin{sideways}
    \makebox[0pt][l]{\hspace{-3.75cm}
    \begin{minipage}{4cm}
    {\bf \scriptsize SHRED-ROM WITH} \\   {\bf \scriptsize SPHERICAL HARMONICS} \end{minipage}}
    \end{sideways}\subfloat{
        \includegraphics[width=0.19\textwidth]{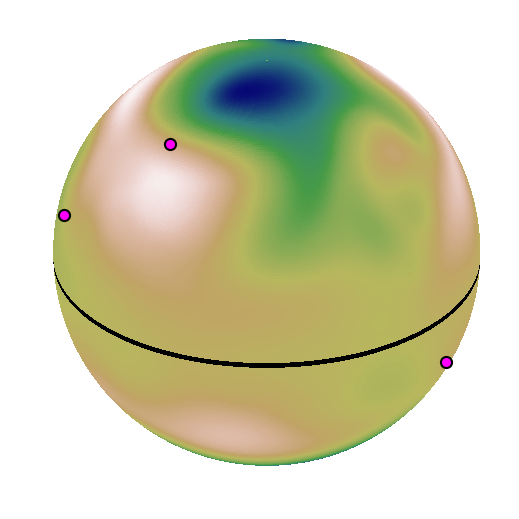}
      
    }\quad
    \subfloat{
        \includegraphics[width=0.19\textwidth]{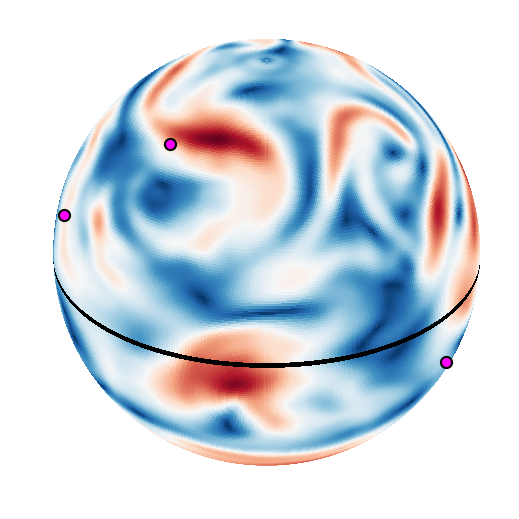}
      
    }\quad
    \subfloat{
        \includegraphics[width=0.19\textwidth]{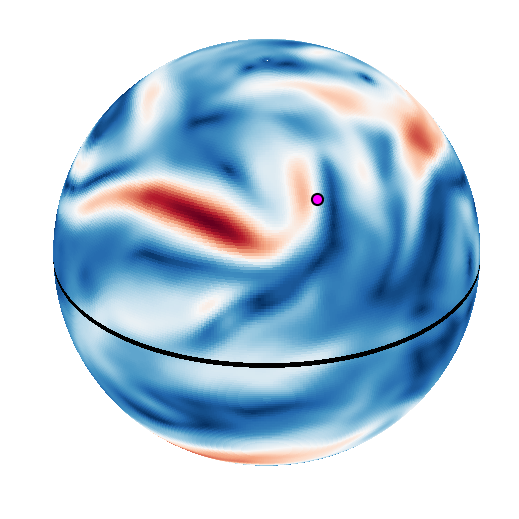}
      
    }\quad
    \subfloat{
        \includegraphics[width=0.19\textwidth]{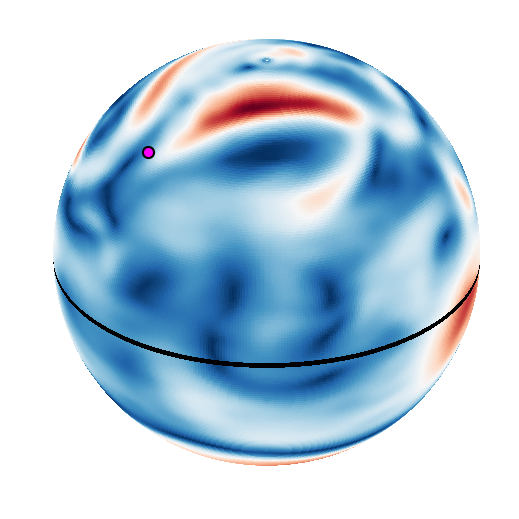}
      
    }
    
    \qquad \subfloat{
        \includegraphics[width=0.42\textwidth]{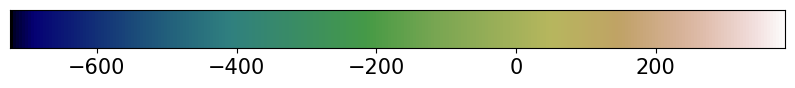}
      
    } \subfloat{
        \includegraphics[width=0.42\textwidth]{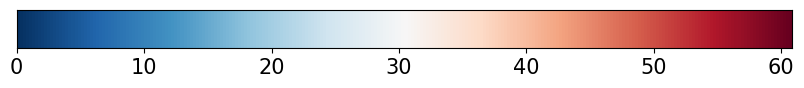}
      
    }

    \caption{{\em Shallow Water Equations}. Ground truth (first row), SHRED-ROM reconstructions with POD-based compressive training (second row) and SHRED-ROM reconstructions with spherical harmonics-based compressive training (third row). The following test cases are considered: height reconstruction from $3$ fixed height sensors (first column); velocity reconstruction from $3$ fixed height sensors (second column); velocity reconstruction from the coordinates of $1$ mobile sensor at $t=11$ hours (third column) and $t=957$ hours (fourth column). The sparse sensors exploited by SHRED-ROM are depicted with magenta dots.}
        \label{fig:SWE}
\end{figure*}

The first test case we consider deal with the reconstruction of synthetic spherical data in a nonparametric setting. The snapshots correspond to the solution of the Shallow Water Equations (SWE) on a sphere with earth-like topography and daily/annual periodic external forces, available in the dataset {\em The Well} ~\cite{ohana2024thewell}. The SWE solution includes three high-dimensional fields ($N_h = 131072$), corresponding to the height deviation of pressure surface from its mean height, and to the velocity components with respect to the polar and azimuthal angles, over a year ($N_t = 1008$ time steps).

We first consider randomized SVD to reduce the data dimensionality and allow for compressive training. With only $r = 50$ POD modes (compression ratio equal to $99.96\%$), it is possible to compress the height snapshots with a relative reconstruction error on test data equal to $1.34 \%$. Instead, $r = 75$ are retrieved for the velocity components reductions (compression ratio equal to $99.94\%$), with relative test errors equal $1.71\%$ and $0.91\%$. Note that the number of POD modes to retrieve for accurate compressions is problem-dependent, and can be determined by looking at the singular values decay.

SHRED-ROM is then applied to reconstruct the high-dimensional spatio-temporal fields starting from $3$ fixed sensors at as many random locations. To show the wide applicability of SHRED-ROM with coupled fields, we suppose that the sensors can monitor the height variable only, without having access to the velocity values. After data preprocessing with a lag $L=100$ and neural networks compressive training, SHRED-ROM can accurately reconstruct the trajectories of the three high-dimensional fields starting from the limited sensors randomly chosen. Specifically, the reconstruction errors on the time instants unseen during training are equal to $3.14 \%$, $9.59 \%$ and $5.90 \%$ for, respectively, the height variable and the polar and azimuthal components of the velocity.

The system configuration may also be captured by mobile sensors, such as drifting buoys. In particular, we consider $1$ sensor passively transported by the velocity field on the sphere. Starting only from the history of the sensor coordinates, thus without monitoring any quantity, SHRED-ROM accurately predicts the velocity components, with test relative errors equal to $7.43 \%$ and $4.80 \%$, respectively.

Instead of considering data-driven compression techniques for the snapshots, such as POD, we can also exploit a different physics-based basis of functions that properly capture the variability of the SWE solution. In this context, we employ the spherical harmonics up to degree $50$, resulting in a dimensionality reduction with $r = 2601$ and with test reconstruction errors equal to $0.25\%$, $9.45\%$ and $6.37\%$ for the height and velocity components, respectively. By measuring the height values in $3$ random locations over time, SHRED-ROM can still accurately predict the spherical harmonics expansion coefficients. The reconstruction errors committed on the test snapshots of height and velocity components are, respectively, $2.15 \%$, $10.71 \%$ and $7.13 \%$. If, instead, we take into account a mobile sensor passively transported by the velocity field, SHRED-ROM reconstructs the high-dimensional polar and azimuthal components of the velocity with test errors equal to $11.64 \%$ and $7.68 \%$. 

Figure~\ref{fig:SWE} shows four height and velocity magnitude snapshots in the test set, along with the corresponding SHRED-ROM reconstructions obtained considering both POD and spherical harmonics as compression techniques. By a visual inspection, it is possible to assess the high level of accuracy of SHRED-ROM despite the limited input information, that are $3$ fixed sensors capturing the height variable or the temporal history of the mobile sensor coordinates. 


\subsection{GoPro Physics}
\label{subsec:GoPro}

\begin{figure*}[t]
    \centering
    \begin{sideways}
    \makebox[0pt][l]{\hspace{-3.35cm} {\bf \footnotesize ORIGINAL VIDEOS}}
    \end{sideways} \subfloat[\shortstack{Symmetric shedding \\ 400th frame}]{
        \includegraphics[width=0.23\textwidth]{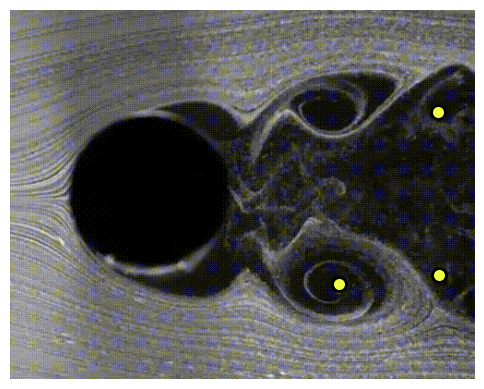}
      
    } 
    \subfloat[\shortstack{Symmetric shedding \\ 800th frame}]{
        \includegraphics[width=0.23\textwidth]{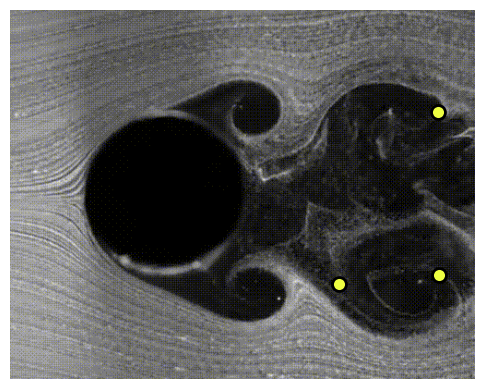}
      
    } 
    \subfloat[\shortstack{Alternating symmetric shedding \\ 269th frame}]{
        \includegraphics[width=0.23\textwidth]{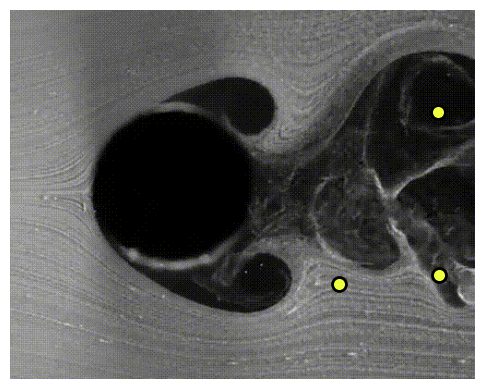}
      
    } 
    \subfloat[\shortstack{Alternating symmetric shedding \\ 1002nd frame}]{
        \includegraphics[width=0.23\textwidth]{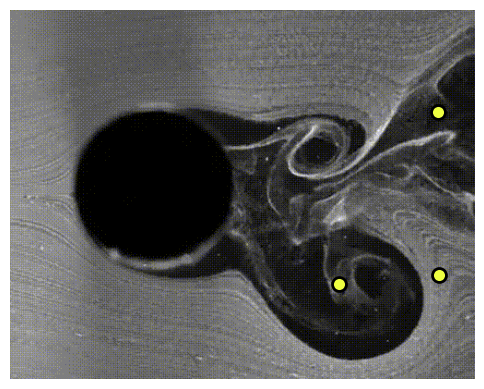}
      
    }

        \begin{sideways}
    \makebox[0pt][l]{\hspace{-2.15cm} {\bf \footnotesize DATA}}
    \end{sideways} \subfloat{
        \includegraphics[width=0.23\textwidth]{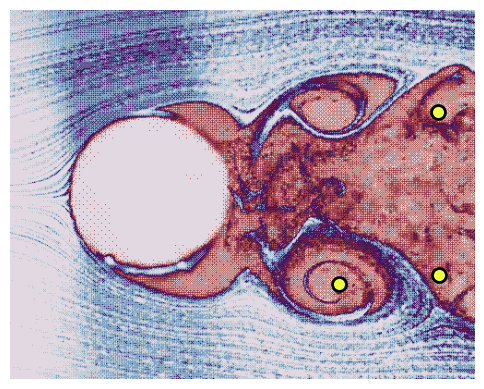}
      
    }
    \subfloat{
        \includegraphics[width=0.23\textwidth]{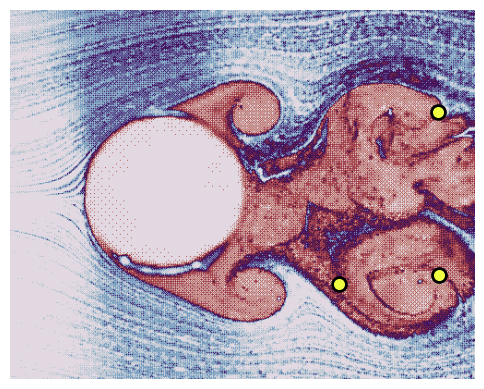}
      
    }
    \subfloat{
        \includegraphics[width=0.23\textwidth]{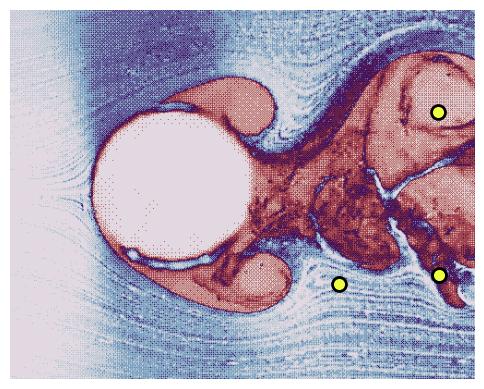}
      
    }
    \subfloat{
        \includegraphics[width=0.23\textwidth]{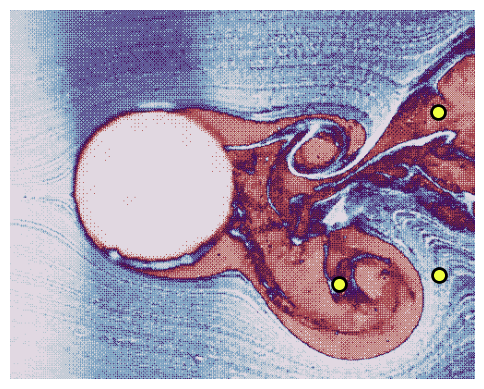}
      
    }

        \begin{sideways}
    \makebox[0pt][l]{\hspace{-2.8cm} {\bf \footnotesize SHRED-ROM}}
    \end{sideways} \subfloat{
        \includegraphics[width=0.23\textwidth]{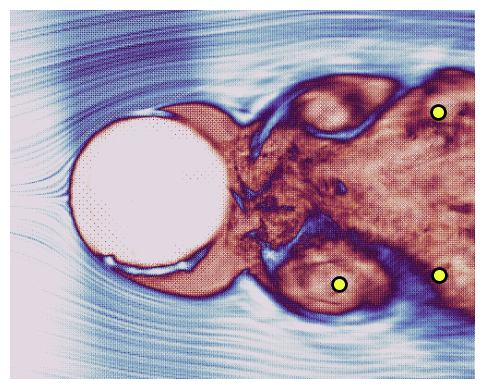}
      
    }
    \subfloat{
        \includegraphics[width=0.23\textwidth]{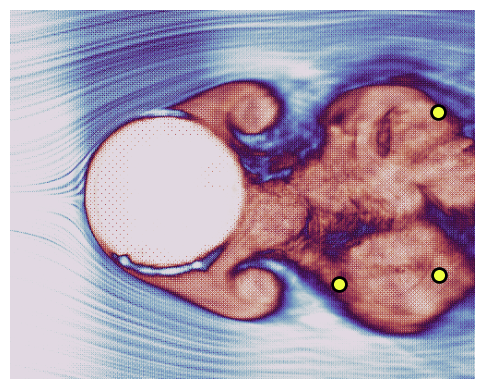}
      
    }
    \subfloat{
        \includegraphics[width=0.23\textwidth]{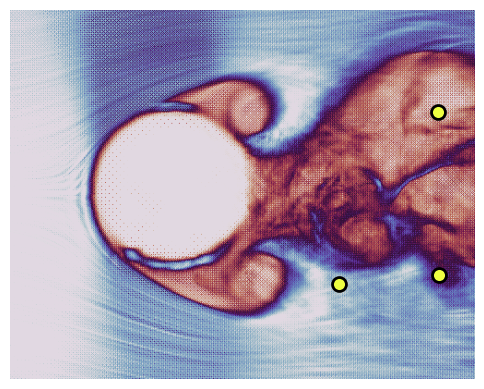}
      
    }
    \subfloat{
        \includegraphics[width=0.23\textwidth]{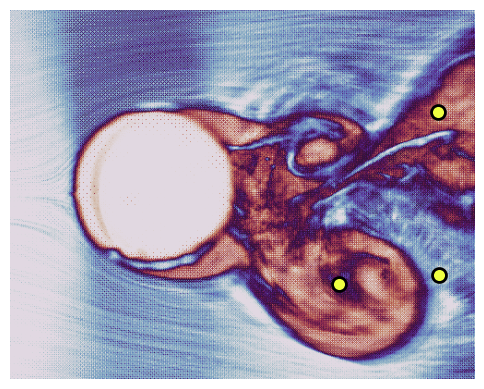}
      
    }
    \vspace{0.2cm}

    \hspace{0.35cm}\subfloat{
        \includegraphics[width=0.43\textwidth]{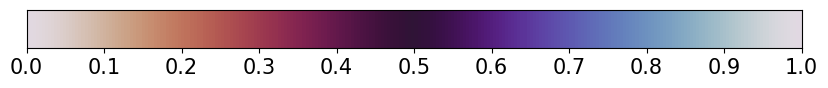}
      
    }
    
    \caption{{\em GoPro Physics}. Original videos (first row), preprocessed data (second row) and SHRED-ROM reconstructions from $3$ fixed pixels in the cylinder wake (first column). The following test cases are considered: symmetric shedding reconstruction of the $400$th (first column) and the $800$th frame (second column);  alternating symmetric shedding reconstruction of the $269$th (third column) and the $1002$nd frame (fourth column). The sparse pixels exploited by SHRED-ROM are depicted with yellow dots.}
        \label{fig:GoPro}
\end{figure*}

The second test case focuses on reconstructing videos of experimental studies in fluid dynamics, typically referred to as {\em GoPro physics}, in contrast with the other examples where we exploit synthetic snapshots simulated through high-fidelity solvers. Specifically, we consider two videos recording the vortex shedding in the wake of an oscillating cylinder provided by Boersma et al.~\cite{vortexarms_videos}. Due to different inflow velocities considered, two different patterns are displayed, that are the so-called {\em symmetric} and {\em alternating symmetric shedding}. Starting from sparse pixels values, we aim to build a SHRED-ROM capable of reconstructing the videos frames, while discriminating between the two scenarios.

Each video consists of $N_t = 1077$ frames of dimension $400 \times 504$ ($N_h = 201600$). In the preprocessing stage, the original frames are converted to gray scale, the background is removed and the pixel values are normalized. POD is then applied to allow for compressive training. Specifically, retaining the first $r=100$ POD coefficients for every frame with randomized SVD (that is a compression ratio equal to $99.95\%$), we achieve a POD reconstruction error equal to  $11.96\%$.

SHRED-ROM is taken into account to reconstruct the frames starting from $3$ pixels randomly sampled within the region behind the cylinder. With a lag parameter equal to $L=150$, SHRED-ROM accurately approximates the test snapshots with a relative error equal to $13.17\%$. In the direction of parametric settings, it is interesting to note that SHRED-ROM is able to automatically discriminate among the two videos from the pixel values, without requiring any knowledge about the underlying setting. Figure~\ref{fig:GoPro} presents $4$ test snapshots of the original videos, along with the corresponding preprocessed data and the SHRED-ROM reconstructions.


\subsection{Kuramoto-SivashinskyEquation}
\label{subsec:KS}

\begin{figure*}[t]
    \centering
    
    \begin{sideways}
    \makebox[0pt][l]{\hspace{-1.87cm} {\bf \footnotesize DATA}}
    \end{sideways}  \subfloat[\shortstack{$\boldsymbol{\mu}=[1.27, 3.71]^{\top}$}]{
    \begin{overpic}[width=0.3\textwidth]{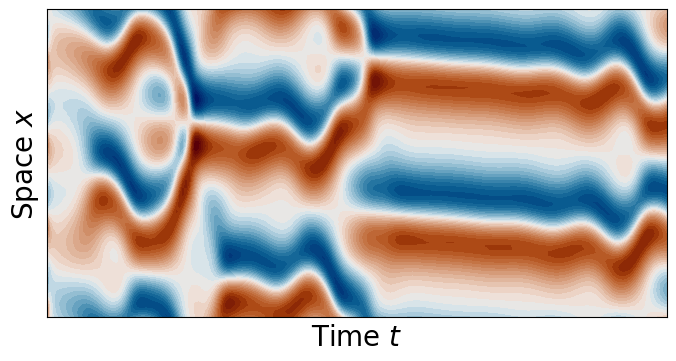}
    \put(-10.5,57){\bf (a)}
    \end{overpic}      
    }
    \subfloat[\shortstack{$\boldsymbol{\mu} = [1.18, 4.34]^{\top}$}]{
        \includegraphics[width=0.3\textwidth]{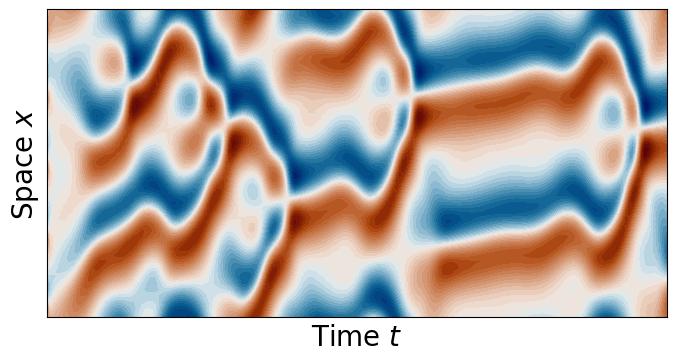}
      
    }
    \subfloat[\shortstack{$\boldsymbol{\mu} = [1.33, 1.35]^{\top}$}]{
        \includegraphics[width=0.3\textwidth]{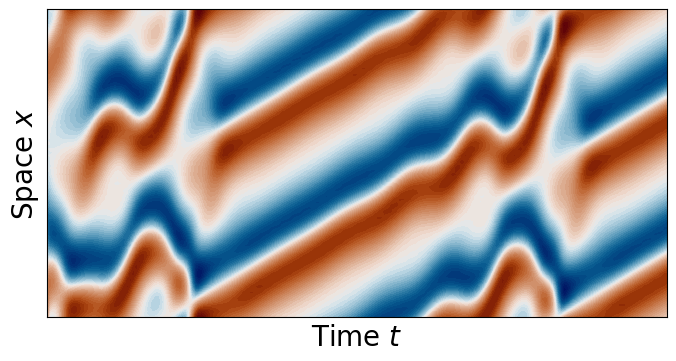}
      
    }

    \begin{sideways}
    \makebox[0pt][l]{\hspace{-2.5cm} {\bf \footnotesize SHRED-ROM}}
    \end{sideways} \subfloat{
        \includegraphics[width=0.3\textwidth]{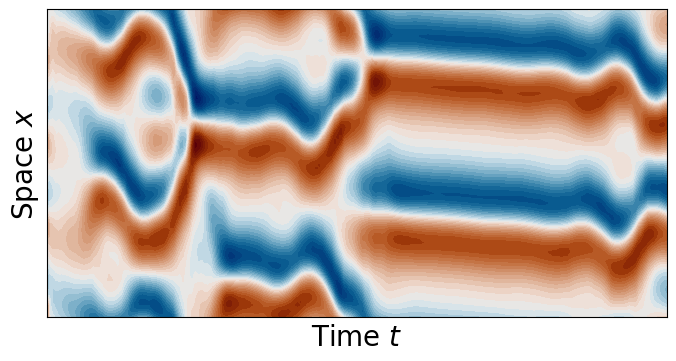}
      
    }
    \subfloat{
        \includegraphics[width=0.3\textwidth]{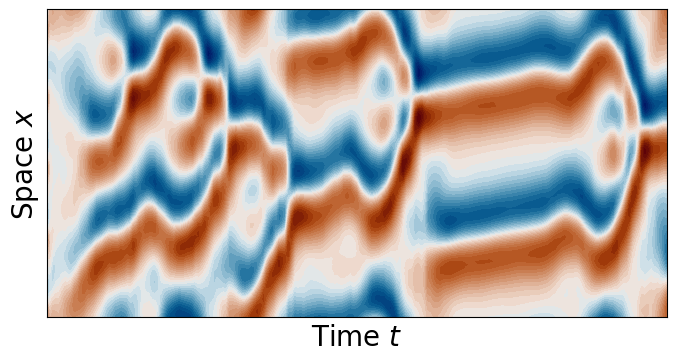}
      
    }
    \subfloat{
        \includegraphics[width=0.3\textwidth]{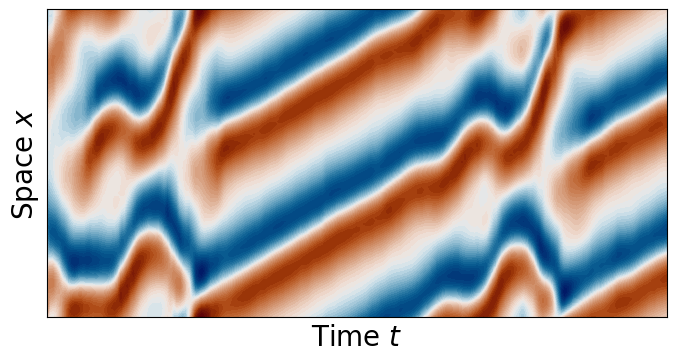}
      
    }
    \vspace{0.2cm}

    \hspace{0.35cm}\subfloat{
        \includegraphics[width=0.4\textwidth]{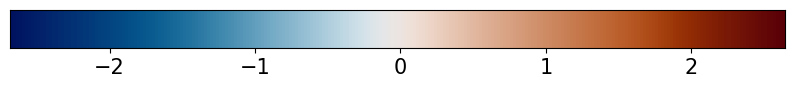}
      
    }
    
    \subfloat{
        \begin{overpic}[width=0.42\textwidth]{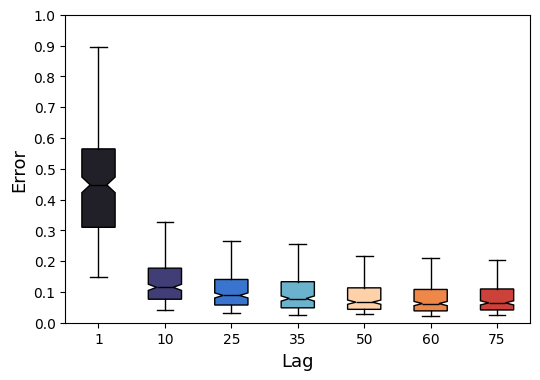}
    \put(-11.5,67){\bf (b)}
    \end{overpic}
    }
    \quad
    \subfloat{
        \includegraphics[width=0.42\textwidth]{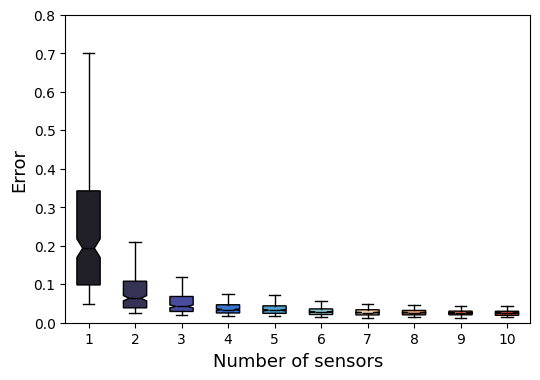}
      
    }
    
    \caption{{\em Kuramoto-Sivashinsky}. {\bf (a)} Ground truth (first row) and SHRED-ROM reconstructions (second row). The following test cases are considered: state reconstruction from $2$ fixed sensors with $\boldsymbol{\mu} = [1.27, 3.71]^{\top}$ (first column), $\boldsymbol{\mu} = [1.18, 4.34]^{\top}$ (second column) and $\boldsymbol{\mu} = [1.33, 1.35]^{\top}$ (third column). {\bf (b)} Relative reconstruction errors committed by SHRED-ROM on test data for different lag values (first column) and different number of sensors (second column), while considering different sensor placements.}
        \label{fig:KS}
\end{figure*}

In this section, we cope with a nonlinear and chaotic fluid dynamics exhibiting parametric dependency. In particular, we model the dynamics of a 1D state variable $u: [0, L] \times [0, T] \to \mathbb{R}$ with the Kuramoto-Sivashinsky equation (KS)
\begin{equation*}
u_t + u_{xx} + \nu u_{xxxx} + u u_x = 0
\end{equation*}
with periodic boundary conditions. To deal with chaotic patterns, we set $L=22$ and $T=200$. Regarding the initial condition, we consider 
\begin{equation*}
u(x,0,\omega) = \cos \left( \frac{2 \pi \omega}{L}x \right) \left(1 + \sin \left( \frac{2 \pi \omega}{L}x \right) \right)\end{equation*}
The viscosity $\nu$ and the frequency of the initial datum are regarded as parameters, that is $\boldsymbol{\mu} = [\nu, \omega]^{\top}$. As discussed in Section~\ref{subsec:SHRED-ROM_model}, our goal is to employ SHRED-ROM to rapidly and accurately approximate the state trajectory for a new set of parameters, unseen in the offline phase. To this aim, we generate $500$ trajectories corresponding to parameters $\nu$ and $\omega$ randomly sampled in the intervals $[1, 2]$ and $[1, 5]$, respectively. Note that an adequate exploration of the parameter space is crucial due to the chaotic and instable patterns given by KS, where small changes in the parameter values result in completely different solutions. To solve the KS equation, we consider the Exponential Time Differencing fourth-order Runge-Kutta (ETDRK4) method~\cite{kassam_ERTDRK4} with a uniform space discretization of $N_h = 100$ spatial points and a time step equal to $\Delta t = 0.01$. Finally, starting from the initial values at $t=0$, we save the KS solution with a frequency of $100$, resulting in trajectories of $N_t = 201$ snapshots. The dimensionality of the generated data is then reduced through randomized SVD: $r=20$ POD modes are sufficient to capture most of the variability (compression ratio equal to $80\%$), with a relative error on test data equal to $0.35\%$.

After generating and compressing the data, it is now possible to train SHRED-ROM. We consider $2$ fixed sensors randomly sampled in the domain and a lag parameter equal to $L=50$. When reconstructing the state trajectories corresponding to new parameters unseen during training, the relative error with respect to the ground truth is equal to $9.13\%$. Note that only the sensor values are exploited as input, disregarding any information about the parameters values -- which, in general, may be unknown or uncertain -- as typically considered by other ROM strategies. The performance of SHRED-ROM can be also visually assessed in Figure~\ref{fig:KS}, panel {\bf (a)}, where we compare three test spatio-temporal KS solutions along with the corresponding SHRED-ROM approximations.

The panel {\bf (b)} of Figure~\ref{fig:KS} presents, instead, an analysis of the SHRED-ROM test reconstruction errors with respect to the lag parameter $L$ (while considering $2$ fixed sensors) and the number of sensors (with lag $L=50$). To show that the proposed architecture is agnostic to sensor placement, every setting takes into account the errors committed by $10$ different SHRED-ROMs trained on as many sensor configurations. The fast decay of the error distributions suggest that even a small temporal history, as well as a tiny amount of sensors, are enough to achieve acceptable results. Note that the case with unitary lag ($L=1$) corresponds to the one-shot reconstruction through a shallow recurrent decoder with compressive training, equivalent to the {\em POD-based deep state estimation} (PDS) proposed by Nair and Goza~\cite{nair2020}.

Sensor measurements in real-world problems are typically corrupted by noise. If we corrupt the sensor values with Gaussian random noise with zero mean and standard deviation equal to $0.25$ (which corresponds approximately to the $5\%$ of the state data range), the generalization capabilities of SHRED-ROM in test scenarios get worse, with a relative reconstruction error equal to $23.24\%$. In this case, thanks to the lightweight SHRED-ROM architecture and the efficient compressive training strategy, we can consider an ensemble of SHRED-ROMs in order to provide a robust state estimation, as proposed by Riva et al.~\cite{riva2024robuststateestimationpartial}. We thus consider $20$ different SHRED-ROMs trained on $2$ sensor trajectories corrupted by as many Gaussian random noises. The state reconstruction can be now obtained by averaging the predictions of every SHRED-ROM, with a mean relative reconstruction errors on test data equal to $15.51 \%$


\subsection{Fluidic Pinball}
\label{subsec:pinball}
\begin{figure*}[t]
    \centering
    \begin{sideways}
    \makebox[0pt][l]{\hspace{-2.5cm} {\bf \footnotesize DATA}}
    \end{sideways} \subfloat[\shortstack{Fixed sensors \\ $\boldsymbol{\mu} = [-1.38, -4.56, -2.54]^{\top}$ \\ 1 second}]{
    \begin{overpic}[width=0.21\textwidth]{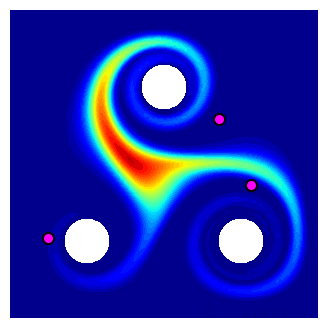}
    \put(-15,125){\bf (a)}
    \end{overpic}
    }
    \subfloat[\shortstack{Fixed sensors \\ $\boldsymbol{\mu} = [-1.38, -4.56, -2.54]^{\top}$ \\ 3 seconds}]{
        \includegraphics[width=0.21\textwidth]{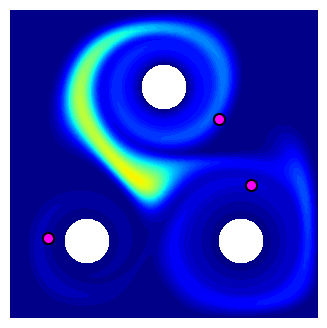}
      
    }
        \subfloat[\shortstack{Mobile sensor \\ $\boldsymbol{\mu} = [-1.42, 4.65, 2.68]^{\top}$ \\ 1 second}]{
        \includegraphics[width=0.21\textwidth]{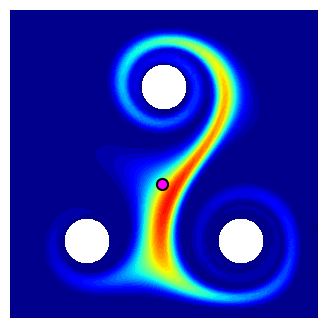}
      
    }
    \subfloat[\shortstack{Mobile sensor \\ $\boldsymbol{\mu} = [-1.42, 4.65, 2.68]^{\top}$ \\ 3 seconds}]{
        \includegraphics[width=0.21\textwidth]{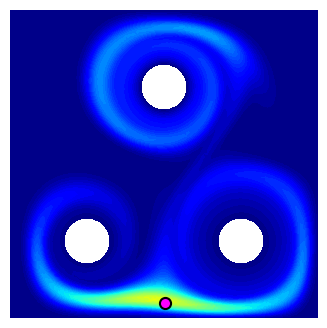}
      
    }

    \begin{sideways}
    \makebox[0pt][l]{\hspace{-2.9cm} {\bf \footnotesize SHRED-ROM}}
    \end{sideways}\subfloat{
        \includegraphics[width=0.21\textwidth]{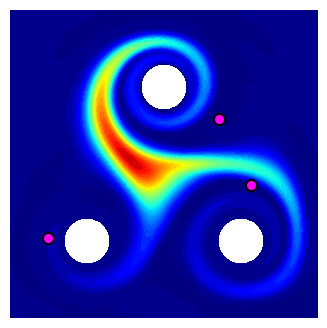}
      
    }
    \subfloat{
        \includegraphics[width=0.21\textwidth]{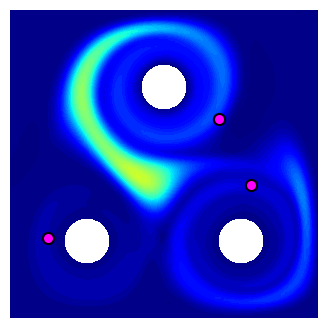}
      
    }
    \subfloat{
        \includegraphics[width=0.21\textwidth]{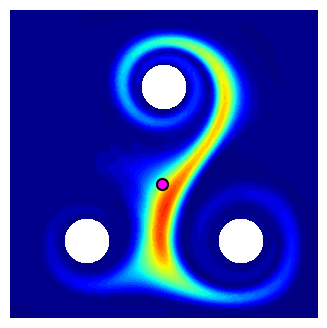}
      
    }
    \subfloat{
        \includegraphics[width=0.21\textwidth]{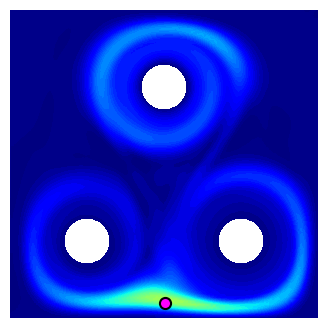}
      
    }
    \vspace{0.2cm}

    \hspace{0.35cm}\subfloat{
        \includegraphics[width=0.4\textwidth]{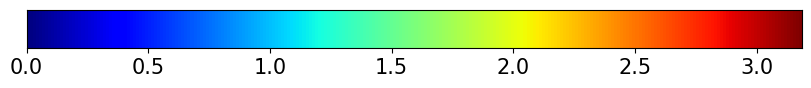}
      
    }

    \subfloat{
    \begin{overpic}[width=0.42\textwidth]{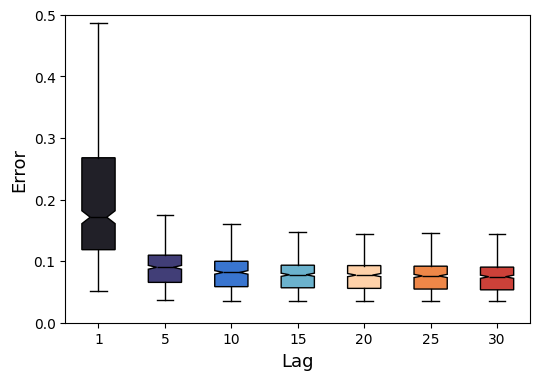}
    \put(-5,67){\bf (b)}
    \end{overpic}       
    }
    \quad
    \subfloat{
        \includegraphics[width=0.42\textwidth]{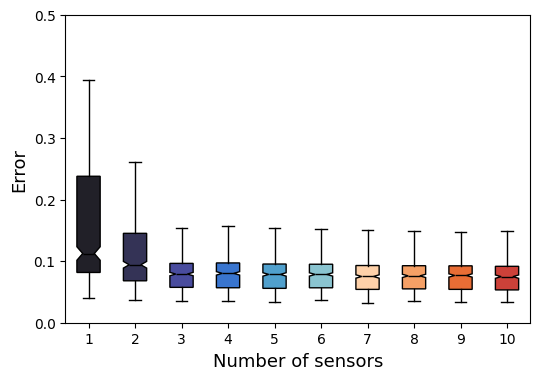}
      
    }
    \caption{{\em Fluidic Pinball}. {\bf (a)} Ground truth (first row) and SHRED-ROM reconstructions (second row). The following test cases are considered: state reconstruction from $3$ fixed sensors with $\boldsymbol{\mu} = [-1.38, -4.56, -2.54]^{\top}$ at $t=1$ seconds (first column) and $t=3$ seconds (second column); state reconstruction from the coordinates of $1$ mobile sensor with $\boldsymbol{\mu} = [-1.42, 4.65, 2.68]^{\top}$ at $t=1$ seconds (third column) and $t=3$ seconds (fourth column). {\bf (b)} Relative reconstruction errors committed by SHRED-ROM on test data for different lag values (first column) and different number of sensors (second column), while considering different sensor placements. The sparse sensors exploited by SHRED-ROM are depicted with magenta dots.}
    \label{fig:pinball}
\end{figure*}

The fourth test case copes with an involved 2D advection-diffusion problem characterized by an implicit parametric dependence. We consider an incompressible fluid constrained in a square domain $(-1,1)^2$ with three cylinders centered at $(-0.5, -0.5)$, $(0.5, -0.5)$ and $(0.0, 0.5)$, and with radius $0.15$. The cylinders can rotate with constant velocities $v_i$ for $i = 1,2,3$ -- here regarded as parameters $\boldsymbol{\mu} = [v_1, v_2, v_3]^{\top}$ -- resulting in a fluid motion inside the box. Specifically, the fluid velocity ${\bf v}: [-1,1]^2 \to \mathbb{R}^2$ and pressure $p: [-1,1]^2 \to \mathbb{R}$ are determined by the steady Navier-Stokes equations
\begin{equation*}
\begin{cases}
- \nu \Delta \mathbf{v} + (\mathbf{v} \cdot \nabla) \mathbf{v} + \nabla p = 0
\\
\nabla \cdot \mathbf{v} = 0
\end{cases}
\end{equation*}
with viscosity $\nu = 1.0$, no-slip boundary conditions on the external walls and Dirichlet boundary conditions on the three cylinders.

Let us now consider a time-dependent quantity $y:$~$[-1,1]^2 \times [0,T] \to \mathbb{R}$ -- which may represent, for instance, heat, mass or density of particles -- that spreads and moves in the domain according to the advection-diffusion PDE
\begin{equation*}
y_t + \nabla \cdot (- \eta \nabla y + \mathbf{v}(\boldsymbol{\mu}) y) = 0
\end{equation*}
with homogeneous Neumann boundary conditions and initial condition
\begin{equation*}
y({\bf x}, 0) = \dfrac{10}{\pi} exp(- 10 x_1^2 - 10 x_2^2).
\end{equation*}
Therefore, starting from a Gaussian function with mean $(0,0)$ and variance $0.05$, the quantity $y$ is transported and deformed by the parametric fluid flow with velocity ${\bf v}(\boldsymbol{\mu})$. To mainly focus on the advection effect over the diffusion one, we set the viscosity $\eta = 0.001$. Moreover, we set the final time $T=3$.

SHRED-ROM is now exploited to provide the state evolution for new combinations of cylinder velocities, unseen during training. To train the LSTM ${\bf f}_T$ and the SDN ${\bf f}_X$, we generate $500$ trajectories for different parameter values in the parameter space $\mathcal{P} = [-5, 5]^3$. Note that both clockwise and counter-clockwise rotations are considered according to the signs of the sampled parameters. Both the steady Navier-Stokes equations and the advection-diffusion PDE are solved with finite element solvers in \texttt{fenics}~\cite{fenics}, taking into account a time step $\Delta t = 0.1$ -- resulting in state trajectories of length $N_t = 31$ -- and a mesh with $N_h = 7525$ vertices. The dimensionality of the simulated snapshots is then reduced to $r=200$ thanks to randomized SVD. Despite a compression ratio equal to $97\%$, the low-dimensional features are enough to accurately reconstruct the high-dimensional data with a test error equal to $1.54\%$.

We now build two SHRED-ROMs to deal with, respectively, fixed and mobile sensors. In the first case, we measure the state $y$ at $3$ fixed locations within the domain. In the second setting, we suppose to know only the spatial coordinates of $1$ mobile sensor located in $(0,0)$ at $t=0$ and passively transported by the fluid flow. Note that, to show the versatility of the method, we suppose to know the parameter values $\boldsymbol{\mu}$, and we thus exploit this information as input, in addition to the sensor values. After training the corresponding LSTM networks and the SDNs with a time window of length $L=10$, both SHRED-ROMs are able to reconstruct the state patterns in new scenarios with mean relative errors equal to, respectively, $7.90\%$ and $8.44\%$. The panel {\bf (a)} of Figure~\ref{fig:pinball} displays four different test snapshots along with the corresponding reconstructions provided by SHRED-ROM, both taking into account fixed and mobile sensors. The panel {\bf (b)} of Figure~\ref{fig:pinball} shows, instead, the behavior of the mean relative error on test data with respect to different lag values and different number of fixed sensors, while taking into account $10$ different random sensor placements in each setting. Similarly to the findings in Section~\ref{subsec:KS}, a small time window $L$ and a very limited amount of sensors are enough to achieve accurate results. In particular we note that, in both the parametric applications presented so far, accurate results are obtained by exploiting $d+1$ fixed sensors, where $d$ is the problem dimension ($d=1$ in the Kuramoto-Sivashinsky setting, $d=2$ in the fluidic pinball one). Similarly, an object in a $d$-dimensional space can be localized by $d+1$ constraints, such as $d+1$ sensors measuring the distance to that object. 


\subsection{Flow Around an Obstacle}
\label{subsec:NS}

\begin{figure*}
    \centering
    \begin{sideways}
    \makebox[0pt][l]{\hspace{-1.3cm}
    \bf \footnotesize DATA}
    \end{sideways}\subfloat[\shortstack{$\boldsymbol{\mu} = [-0.40, 2.21, 0.46, 0.91]^{\top}$ \\ 5 seconds}]{
    \includegraphics[width=0.42\textwidth]{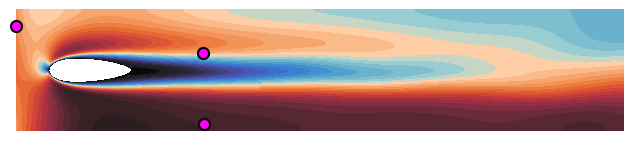}}
    \subfloat[\shortstack{$\boldsymbol{\mu}(t) = [\cos(0.32 t + 3.68), 5.77, 0.45, 0.47]^{\top}$ \\ 5 seconds}]{\includegraphics[width=0.42\textwidth]{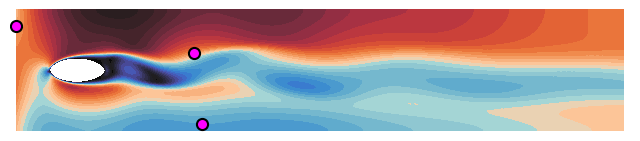}}

    \begin{sideways}
    \makebox[0pt][l]{\hspace{-2cm}
    \bf \footnotesize SHRED-ROM}
    \end{sideways}\subfloat{
    \includegraphics[width=0.42\textwidth]{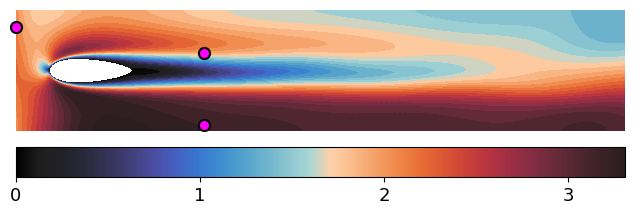}}
    \subfloat{\includegraphics[width=0.42\textwidth]{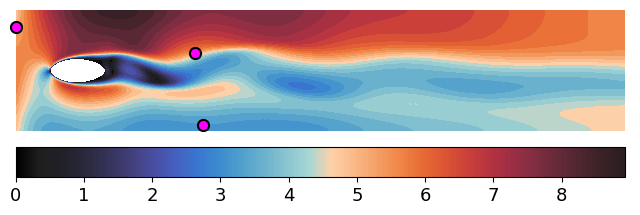}}
    \vspace{0.2cm}

    \hspace{-0.56cm}
    \begin{sideways}
    \makebox[0pt][l]{\hspace{-4.7cm}
    \bf \footnotesize PARAMETER ESTIMATION}
    \end{sideways}\hspace{0.1cm}\subfloat{
    \includegraphics[width=0.38\textwidth]{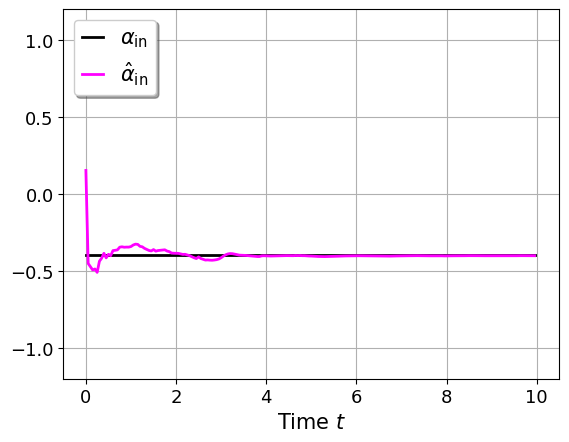}}\qquad \;
    \subfloat{\includegraphics[width=0.38\textwidth]{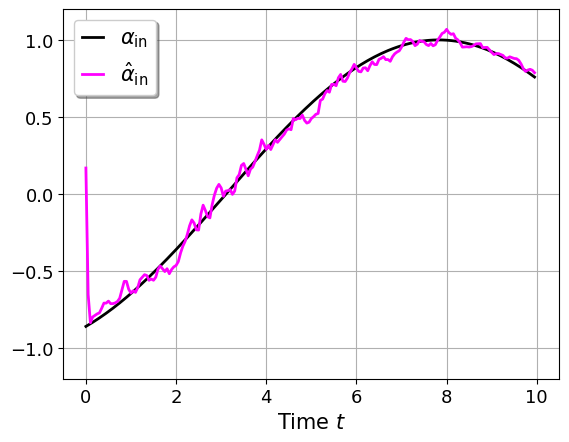}}
    
    \caption{{\em Fluid Around an Obstacle}. Ground truth (first row), SHRED-ROM reconstructions (second row) and parameter estimations (third row). The following test cases are considered: state reconstruction at $t=5$ seconds and parameter estimation over $[0,10]$ seconds from $3$ fixed horizontal velocity sensors with $\boldsymbol{\mu} = [-0.40, 2.21, 0.46, 0.91]^{\top}$ (first column) and $\boldsymbol{\mu}(t) = [\cos(0.32 t + 3.68), 5.77, 0.45, 0.47]^{\top}$ (second column). The sparse sensors exploited by SHRED-ROM are depicted with magenta dots.}
    \label{fig:NS}
\end{figure*}

\begin{figure*}
    \centering
    \begin{sideways}
    \makebox[0pt][l]{\hspace{-1.1cm}
        \begin{rotate}{-45}
            \begin{minipage}{1cm}
                {\bf \scriptsize DATA}
            \end{minipage}
        \end{rotate}}
        \end{sideways} \hspace{0.5cm} \subfloat[\shortstack{$\boldsymbol{\mu}(t)=[\cos(2.2t + 2.3), 5.19, 0.22, 0.53]^{\top}$ \\ 2.5 seconds}]{
        \begin{overpic}[width=0.31\textwidth]{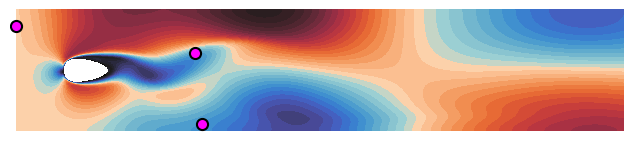}
        \put(-14,33){\bf (a)}
        \end{overpic} 
        } \hspace{-0.5cm}
    \subfloat[\shortstack{$\boldsymbol{\mu}(t)=[\cos(2.2t + 2.3), 5.19, 0.22, 0.53]^{\top}$ \\ 5.0 seconds}]{
        \includegraphics[width=0.31\textwidth]{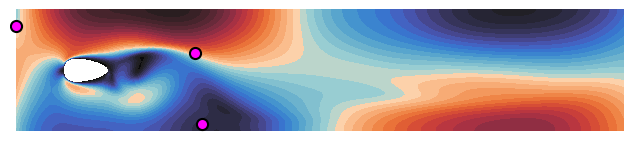}
      
    } \hspace{-0.5cm}
    \subfloat[\shortstack{$\boldsymbol{\mu}(t)=[\cos(2.2t + 2.3), 5.19, 0.22, 0.53]^{\top}$ \\ 7.5 seconds}]{
        \includegraphics[width=0.31\textwidth]{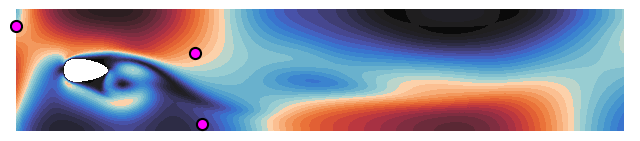}
      
    }
    
    \begin{sideways}
    \makebox[0pt][l]{\hspace{-1.2cm}
        \begin{rotate}{-45}
            \begin{minipage}{1cm}
                {\bf \scriptsize SHRED-ROM}
            \end{minipage}
        \end{rotate}}
        \end{sideways} \hspace{0.5cm} \subfloat{
        \includegraphics[width=0.31\textwidth]{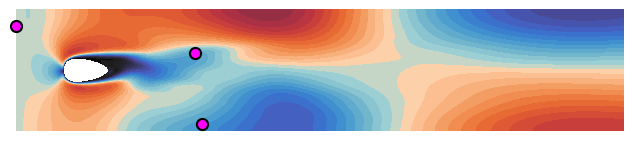}
      
    } \hspace{-0.5cm}
    \subfloat{
        \includegraphics[width=0.31\textwidth]{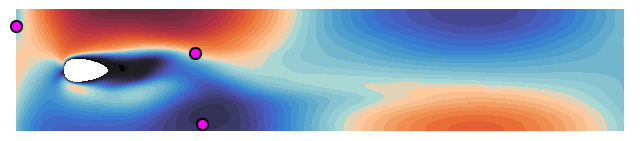}
      
    } \hspace{-0.5cm}
    \subfloat{
        \includegraphics[width=0.31\textwidth]{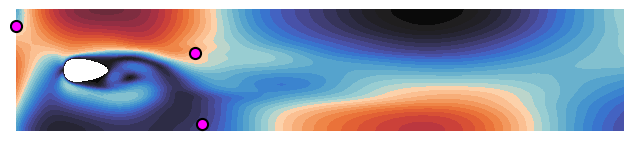}
      
    }

    \begin{sideways}
    \makebox[0pt][l]{\hspace{-1.2cm}
        \begin{rotate}{-45}
            \begin{minipage}{1cm}
                {\bf \scriptsize PDS}
            \end{minipage}
        \end{rotate}}
        \end{sideways} \hspace{0.5cm} \subfloat{
        \includegraphics[width=0.31\textwidth]{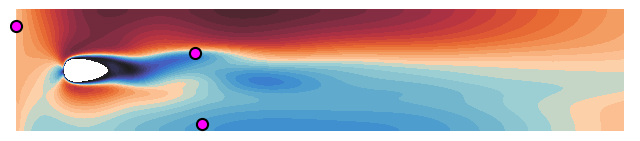}
      
    } \hspace{-0.5cm}
    \subfloat{
        \includegraphics[width=0.31\textwidth]{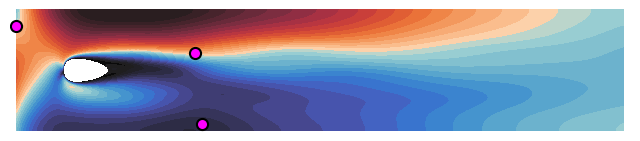}
      
    } \hspace{-0.5cm}
    \subfloat{
        \includegraphics[width=0.31\textwidth]{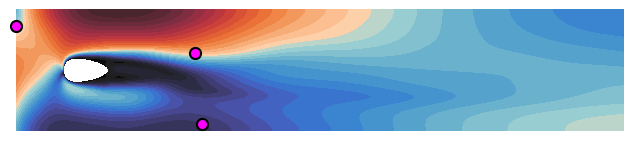}
      
    }

    \begin{sideways}
    \makebox[0pt][l]{\hspace{-1.2cm}
        \begin{rotate}{-45}
            \begin{minipage}{1cm}
                {\bf \scriptsize POD-AE-SE}
            \end{minipage}
        \end{rotate}}
        \end{sideways} \hspace{0.5cm} \subfloat{
        \includegraphics[width=0.31\textwidth]{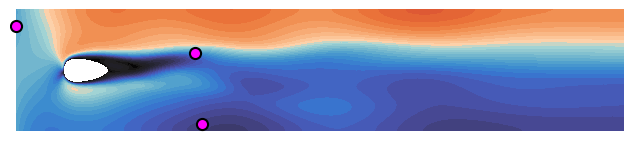}
      
    } \hspace{-0.5cm}
    \subfloat{
        \includegraphics[width=0.31\textwidth]{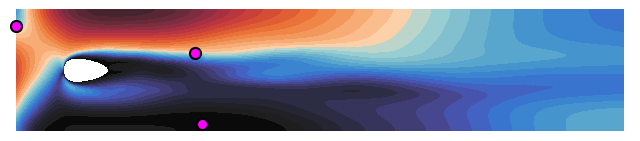}
      
    } \hspace{-0.5cm}
    \subfloat{
        \includegraphics[width=0.31\textwidth]{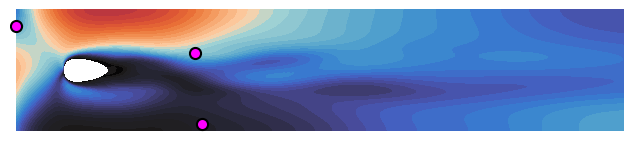}
      
    }

    \begin{sideways}
    \makebox[0pt][l]{\hspace{-1.2cm}
        \begin{rotate}{-45}
            \begin{minipage}{1cm}
                {\bf \scriptsize POD-DeepO Net}
            \end{minipage}
        \end{rotate}}
        \end{sideways} \hspace{0.5cm} \subfloat{
        \includegraphics[width=0.31\textwidth]{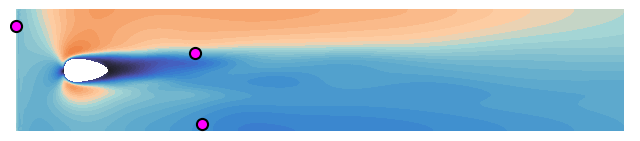}
      
    } \hspace{-0.5cm}
    \subfloat{
        \includegraphics[width=0.31\textwidth]{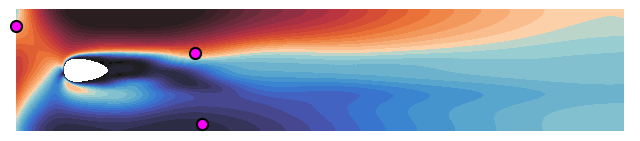}
      
    } \hspace{-0.5cm}
    \subfloat{
        \includegraphics[width=0.31\textwidth]{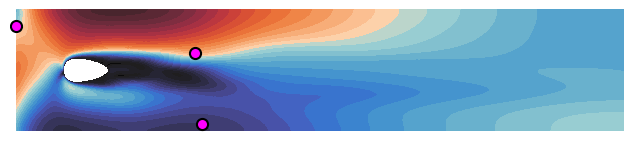}
      
    }

        \begin{sideways}
    \makebox[0pt][l]{\hspace{-1.2cm}
        \begin{rotate}{-45}
            \begin{minipage}{1cm}
                {\bf \scriptsize POD-NN}
            \end{minipage}
        \end{rotate}}
        \end{sideways} \hspace{0.5cm} \subfloat{
        \includegraphics[width=0.31\textwidth]{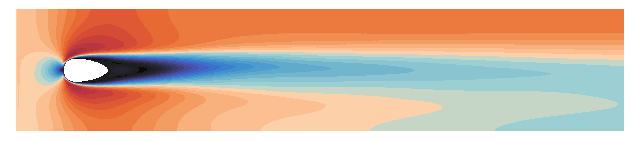}
      
    } \hspace{-0.5cm}
    \subfloat{
        \includegraphics[width=0.31\textwidth]{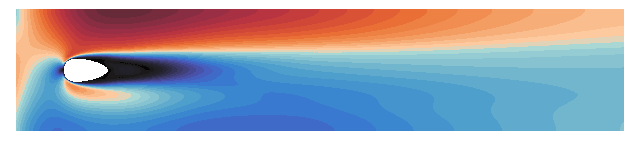}
      
    } \hspace{-0.5cm}
    \subfloat{
        \includegraphics[width=0.31\textwidth]{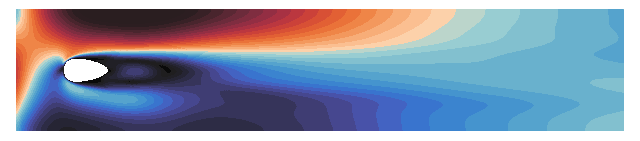}

    }

    \begin{sideways}
    \makebox[0pt][l]{\hspace{-1.2cm}
        \begin{rotate}{-45}
            \begin{minipage}{1cm}
                {\bf \scriptsize POD-DL-ROM}
            \end{minipage}
        \end{rotate}}
        \end{sideways} \hspace{0.5cm} \subfloat{
        \includegraphics[width=0.31\textwidth]{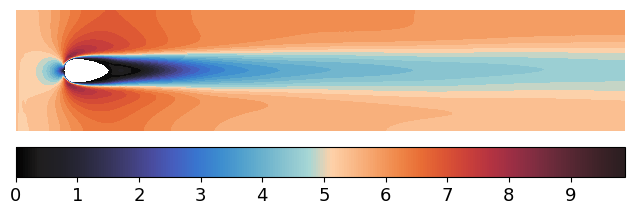}
      
    } \hspace{-0.5cm}
    \subfloat{
        \includegraphics[width=0.31\textwidth]{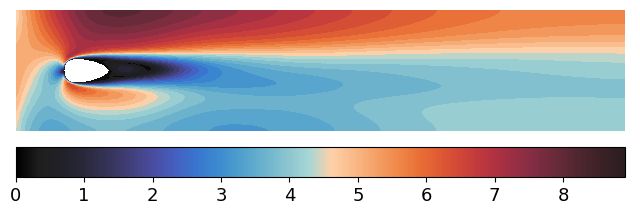}
      
    } \hspace{-0.5cm}
    \subfloat{
        \includegraphics[width=0.31\textwidth]{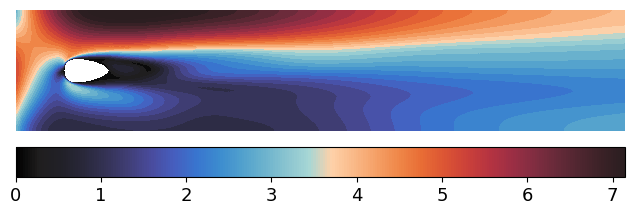}
      
    }
    \vspace{0.2cm}

    \hspace{0.5cm} \subfloat{
    \begin{overpic}[width=0.35\textwidth]{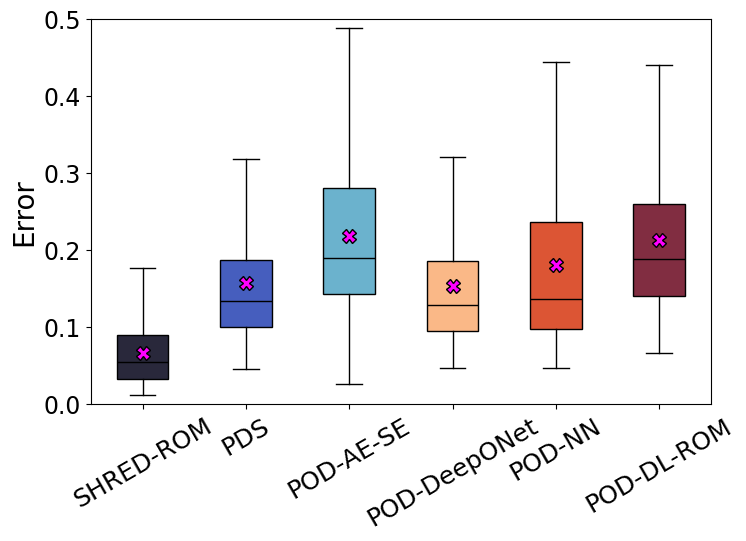}
        \put(-9,68){\bf (b)}
        \end{overpic} 
    }\quad
    \subfloat{
    \begin{overpic}[width=0.55\textwidth]{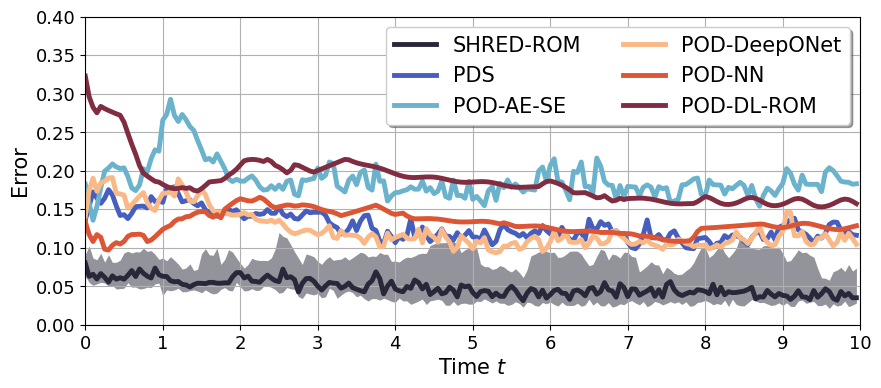}
        \put(-4,41){\bf (c)}
        \end{overpic}
    }

    \hspace{0.5cm}
    \subfloat{
    \begin{overpic}[width=0.35\textwidth]{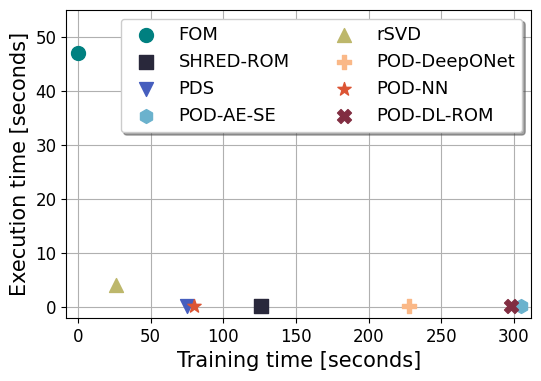}
        \put(-8,65){\bf (d)}
        \end{overpic}
    }
    \quad
    \subfloat{
    \begin{overpic}[width=0.55\textwidth]{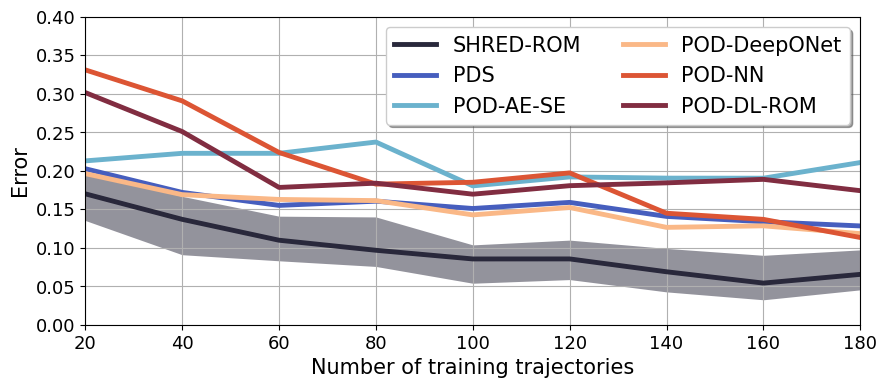}
        \put(-3.5,40){\bf (e)}
        \end{overpic}
    }
    \caption{{\em Fluid Around an Obstacle}. {\bf (a)} Ground truth and competing methods velocity reconstructions with test parameters $\boldsymbol{\mu}(t)=[\cos(2.2t + 2.3), 5.19, 0.22, 0.53]^{\top}$ at $t=2.5$ (first column), $t=5.0$ (second column) and $t=7.5$ seconds (third column). {\bf (b)} Distributions of relative reconstruction errors committed on test data by competing methods. {\bf (c)} Median relative reconstruction errors committed on test data by competing methods over the time horizon $[0,T]$. For SHRED-ROM, also the corresponding interquartile range is displayed. {\bf (d)} Training and execution times required by the {\em full-order model} (FOM), randomized SVD (rSVD) and competing methods. {\bf (e)} Median relative reconstruction errors committed on test data by competing methods for different training set dimensions. For SHRED-ROM, also the corresponding interquartile range is displayed.}
        \label{fig:NS_comparisons}
\end{figure*}

The last test case is devoted to the reconstruction of fluid flows where both time-dependent, physical and geometrical parametric dependencies are employed. Let us consider a 2D channel $[0,10]\times[0,2]$ with a circular obstacle centered in $(1,1)$ and with radius equal to $0.2$. Through Radial Basis Function (RBF) interpolation, it is possible to deform the reference setting in order to obtain different obstacle shapes, without changing the number of degrees of freedom $N_h$. In particular, we lengthen the circle to the left and right by, respectively, $\gamma_l$ and $\gamma_r$, so that the surface of the deformed obstacle passes through the points $(0.8 - \gamma_l, 1.0)$, $(1.0, 1.2)$, $(1.2 + \gamma_r, 1.0)$ and $(1.0, 0.8)$. Let us also consider an incompressible fluid flow around the obstacle, whose dynamics is described by the unsteady Navier-Stokes equations
\begin{equation*}
\begin{cases}
{\bf v}_t - \nu \Delta \mathbf{v} + (\mathbf{v} \cdot \nabla) \mathbf{v} + \nabla p = 0
\\
\nabla \cdot \mathbf{v} = 0
\end{cases}
\end{equation*}
in terms of the velocity ${\bf v}:\Omega(\gamma_l,\gamma_r) \times [0,T] \to \mathbb{R}^2$ and pressure $p:\Omega(\gamma_l,\gamma_r) \times [0,T] \to \mathbb{R}$, where $\Omega(\gamma_l,\gamma_r)$ stands for the spatial parametric domain. Starting from a zero velocity at time $t=0$, the fluid enters the domain from the left boundary with angle of attack $\alpha_{\text{in}}$ and intensity $\gamma_{\text{in}}$, that is
\begin{equation}
{\bf v}((0,x_2),t) = (\gamma_{\text{in}} \cos(\alpha_{\text{in}}), x_2(2-x_2)\gamma_{\text{in}} \sin(\alpha_{\text{in}}))
\label{eq:unsteadyNS}
\end{equation}
where the parabolic profile $x_2(2-x_2)$ is useful to prevent discontinuities. Moreover, free-slip and no-slip boundary conditions are employed to, respectively, the walls and the obstacle.

Our goal is to reconstruct the high-dimensional velocity field over time for different sets of parameters $\boldsymbol{\mu} = [\alpha_{\text{in}}, \gamma_{\text{in}}, \gamma_l, \gamma_r]^{\top}$. To this aim, we generate $200$ velocity trajectories by solving (\ref{eq:unsteadyNS}) through the incremental Chorin-Temam projection method in \texttt{fenics}~\cite{fenics} with $T=10$ and $\Delta t = 0.05$ ($N_t = 200$ time steps). The first $100$ scenarios consider constant parameters randomly sampled in the space $\mathcal{P}=[-1.0,1.0] \times [1.0,10.0] \times [0.2, 0.6] \times [0.2, 1.0]$. The other $100$ velocity trajectories take into account as many time-dependent angles of attack $\alpha_{\text{in}}(t) = \cos\left(\frac{2 \pi}{\tau} t + \vartheta \right)$ with $\tau$ and $\vartheta$ randomly sampled in $(2.5, 20.0)$ and $(0.0, 2 \pi)$, respectively. The dimensionality of the simulated snapshots ($N_h = 80592$) is then reduced to $r=150$ through POD. Despite a remarkably high compression ratio ($99,81\%$), the latent representations are enough to recover the high-dimensional velocities up to a test relative error equal to $1.16\%$.

The spatio-temporal velocity behavior in multiple scenarios can be approximated by SHRED-ROM starting from $3$ fixed sensors monitoring the horizontal velocity only. Besides the POD coefficients of the velocity, we consider the (possibly time-dependent) angle of attack as an additional output of the proposed model, allowing for a real-time parameter estimation through SHRED-ROM. After training the neural networks with a lag equal to $L=50$, SHRED-ROM is capable of reconstructing the high-dimensional velocities in new scenarios with a mean relative error equal to $6.65\%$, and to estimate the corresponding angle of attack with a mean absolute error equal to $0.037$ radians. Figure~\ref{fig:NS} shows the results obtained in two test cases: by a visual inspection it is possible to assess the high accuracy of SHRED-ROM for both the state reconstructions and the parameter estimations. Note that the estimate of the angle of attack $\hat{\alpha}_{\text{in}}$ is not accurate in the first few time steps. However, a very limited temporal history of sensor values allows the estimation to align with respect to the ground truth $\alpha_{\text{in}}$.

With a different training-validation-test splitting, we can estimate state and angle of attack for future times with respect to the ones considered during training. In particular, we now exploit $160$ trajectories in the time range $[0,5]$ for randomized SVD and SHRED-ROM training, while the remaining data (that are the same $160$ trajectories for $5<t\leq10$ and $40$ full trajectories over $[0,10]$) are used for evaluation purposes only. SHRED-ROM is now able to reconstruct the high-dimensional velocity both in new scenarios and future time instants with a mean relative error equal to $7.96\%$. Similar results are obtained for the angle of attack estimation, with a mean absolute error equal to $0.045$ radians.

The SHRED-ROM performances are compared to $5$ different state-of-the-art frameworks arising from different fields. In particular, we take into account {\em (i)} sensor-based state estimation techniques such as {\em POD-based deep state estimation} (PDS)~\cite{nair2020} and {\em POD-enhanced autoencoder state estimation} (POD-AE-SE)~\cite{luo2023}, {\em (ii)} the POD-DeepONet~\cite{lulu2022} operator learning strategy, and {\em (iii)} non-intrusive reduced order modeling techniques such as POD-NN~\cite{hesthaven2018} and {\em POD-enhanced deep learning-based reduced order model} (POD-DL-ROM)~\cite{fresca2022}. Note that, differently from SHRED-ROM, the first three frameworks rely on a one-shot state reconstruction from sensor data, while the last two learn the parameters-to-state map. To perform fair comparisons, every setting takes into account POD-based compressive training with $r=150$ modes, $3$ fixed sensors measuring the horizontal velocity and neural networks with comparable numbers of parameters with respect to SHRED-ROM. For instance, the mappings from sensors or parameters to reduced states show similar complexities with respect to the proposed LSTM, while the encoders and decoders exploited by POD-AE-SE and POD-DL-ROM share the same architecture of the proposed SDN, with a hidden dimension equal to $11$ as suggested by Franco et al.~\cite{Franco2023}. Panel {\bf (a)} of Figure~\ref{fig:NS_comparisons} displays the velocity fields predicted by the aforementioned strategies for $\boldsymbol{\mu}(t)=[\cos(2.2t + 2.3), 5.19, 0.22, 0.53]^{\top}$ at three different time instants. The temporal history of sensor values encoded through the LSTM in SHRED-ROM is crucial to achieve more accurate reconstructions with respect to the competing methods. The superiority of SHRED-ROM is also confirmed by the distributions of the relative test errors shown in panels {\bf (b)} and {\bf (c)} of the same figure. Panel {\bf (d)} compares, instead, the training and execution times of the different frameworks, along with the computational times of the {\em full-order model} (FOM) and randomized SVD (rSVD). While the evaluation times of all the competing methods are extremely fast (ranging between $0.15$ and $0.3$ seconds), with a remarkably high speed up with respect to the FOM, SHRED-ROM is faster to train than POD-DeepONet and autoencoder-based techniques, such as POD-AE-SE and POD-DL-ROM. Note that the training time differences would be even higher whenever non-compressive training strategies were considered. Finally, panel {\bf (e)} of Figure~\ref{fig:NS_comparisons} presents a data requirement analysis across all the competing models. Even with few training trajectories, SHRED-ROM is able to achieve impressive generalization capabilities, superior to the other competitors investigated.


\section{CONCLUSIONS}

In this work, we advocate a new decoding-only reduced order modeling framework to reconstruct high-dimensional fields from temporal history of sparse sensor measurements in multiple scenarios. Importantly, state snapshots reduction through, e.g., POD or spherical harmonics expansion, allows for lightweight neural networks and efficient compressive training strategies, feasible with laptop-level computing. Moreover, differently from other deep learning-based models, minimal hyperparameter tuning is required, as demonstrated throughout the test cases detailed in Section~\ref{sec:test}.

With several applications dealing with chaotic and nonlinear fluid dynamics, as well as video data, we show that the temporal history of at most $3$ sensors is enough to achieve accurate reconstructions for new scenario parameters, with a remarkably high speed up with respect to numerical simulations. Moreover, we demonstrate that SHRED-ROM serves as a versatile framework capable of dealing with fixed and mobile sensors, synthetic and video data, noisy measurements, coupled fields, time extrapolation in periodic regimes, physical and geometric (possibly time-dependent) parametric dependencies, while being agnostic to sensor placement and parameter values. Indeed, differently from traditional ROMs strategies, the parameter values are not required both at training and evaluation stage, allowing for unknown or uncertain parametric dependencies. As shown in Section~\ref{subsec:NS}, SHRED-ROM outperforms competing sensing methods focusing on one-shot reconstructions, as well as non-intrusive ROMs, both in terms of accuracy, training time and data requirement.

The proposed SHRED-ROM framework may be extended in future works in multiple directions. For instance, forecasting may be easily taken into account, as proposed by Williams et al.~\cite{williams2024}, in order to reconstruct the state variables in multiple scenarios even in the absence of sensor values. To this aim, different deep learning-based recurrence strategies may be considered to embed temporal sensor measurements, with accurate forecasts in autoregressive or free-running mode. Moreover, interpretable and sparse dynamical models can be identified at latent level through a SINDy-based regularization, as proposed by Gao et al.~\cite{gao2025sparseidentificationnonlineardynamics} in nonparametric settings, promoting smooth predictions and stable forecasts. Finally, a variational SHRED-ROM may be taken into account to quantify uncertainties in the model reconstructions and in the parameter estimates.
\\ 

\section*{DATA AND CODE AVAILABILITY} 
The data and the code can be found in our repositories \\
\url{https://doi.org/10.5281/zenodo.14524524} \\
\url{https://github.com/MatteoTomasetto/SHRED-ROM}.
\\

\section*{ACKNOWLEDGEMENTS} The work of JNK was supported in part by the US National Science Foundation (NSF) AI Institute for Dynamical Systems (dynamicsai.org), grant 2112085. AM acknowledges the Project “Reduced Order Modeling and Deep Learning for the real- time approximation of PDEs (DREAM)” (Starting Grant No. FIS00003154), funded by the Italian Science Fund (FIS) - Ministero dell'Università e della Ricerca and the project FAIR (Future Artificial Intelligence Research), funded by the NextGenerationEU program within the PNRR-PE-AI scheme (M4C2, Investment 1.3, Line on Artificial Intelligence).

\vfill
\onecolumngrid
\vspace{5mm}
\noindent\centering\rule{0.5\textwidth}{1pt}
\vspace{5mm}
\twocolumngrid
\renewcommand{\bibsection}{}
\bibliographystyle{abbrv}
\bibliography{bibliography,references,merged}

\end{document}